\documentclass[10pt,journal,twocolumn]{IEEEtran}
\ifCLASSOPTIONcompsoc
  \usepackage[nocompress]{cite}
\else
  \usepackage{cite}
\fi

\ifCLASSINFOpdf
\else
\fi

\usepackage{times}
\usepackage{epsfig}
\usepackage{graphicx}
\usepackage{amsmath}
\usepackage{amssymb}
\usepackage{eucal,bibspacing}
\usepackage{cite}
\usepackage{cs}
\usepackage{mathsymb}
\usepackage{textcomp}
\usepackage{verbatim}
\usepackage{subfigure}
\usepackage{url}
\usepackage{multirow}
\usepackage[super]{nth}
\usepackage{colortbl}
\usepackage{pbox}

\usepackage{algorithm2e}
\usepackage{wrapfig}
\usepackage{caption}

\let\savedalgorithm\algorithm
\let\savedendalgorithm\endalgorithm

\usepackage{algorithm}

\renewcommand{\etal}{\textit{et al. }}
\newcommand{\eg}{\textit{e.g., }}
\newcommand{\ie}{\textit{i.e., }}

\begin{document}

\title{Where-and-When to Look: Deep Siamese Attention Networks for Video-based Person Re-identification}

\author{Lin Wu, Yang Wang, Junbin Gao, Xue Li
\IEEEcompsocitemizethanks{\IEEEcompsocthanksitem Lin Wu and Xue Li are with The University of Queensland, St Lucia 4072, Australia; Email: lin.wu@uq.edu.au; xueli@itee.uq.edu.au.

Yang Wang* (Corresponding author) is with Dalian University of Technology, China; Email: yang.wang@dlut.edu.cn.

Junbin Gao is with Discipline of Business Analytics, The University of Sydney Business School, The University of Sydney, NSW 2006, Australia; Email: junbin.gao@sydney.edu.au.
\protect\\
}
}

\IEEEtitleabstractindextext{%
\begin{abstract}
Video-based person re-identification (re-id) is a central application in surveillance systems with significant concern in security. Matching persons across disjoint camera views in their video fragments is inherently challenging due to the large visual variations and uncontrolled frame rates. There are two steps crucial to person re-id, namely discriminative feature learning and metric learning. However, existing approaches consider the two steps independently, and they do not make full use of the temporal and spatial information in videos. In this paper, we propose a Siamese attention architecture that jointly learns spatiotemporal video representations and their similarity metrics. The network extracts local convolutional features from regions of each frame, and enhance their discriminative capability by focusing on distinct regions when measuring the similarity with another pedestrian video. The attention mechanism is embedded into spatial gated recurrent units to selectively propagate relevant features and memorize their spatial dependencies through the network. The model essentially learns which parts (\emph{where}) from which frames (\emph{when}) are relevant and distinctive for matching persons and attaches higher importance therein. The proposed Siamese model is end-to-end trainable to jointly learn comparable hidden representations for paired pedestrian videos and their similarity value. Extensive experiments on three benchmark datasets show the effectiveness of each component of the proposed deep network while outperforming state-of-the-art methods.
\end{abstract}

\begin{IEEEkeywords}
Video-based person re-identification, Gated recurrent units, Spatial correlations, Visual attention.
\end{IEEEkeywords}}

\maketitle

\IEEEdisplaynontitleabstractindextext

\IEEEpeerreviewmaketitle
\section{Introduction}\label{sec:intro}

The person re-identification (re-id) research aims to recognize an individual over disjoint camera views \cite{MidLevelFilter,Xiong2014Person,eSDC,Pedagadi2013Local,Zhao2013SalMatch,YangwangCVIU2018,Wu-PR-Adaptive,Wu-PR-what-where}. It is a problem of practical importance to multimedia and computer vision communities because of its wide range of potential applications, such as the public security and forensic investigation. Also, person re-id is especially beneficial to multimedia content analysis. Examples include human movement analysis \cite{Human-Movement}, clothing retrieval and recommendation \cite{Clothes-parsing,Clothing}, fine-grained object recognition \cite{Diversity-attention,Yang-TIP17,YangCycle2018,YangwangNN2018,YangIJCAI2016,Yang-TIP15,Yangwang2018}, and so on. This is an inherently challenging problem because people appearance are intrinsically limited. The limitations of visual appearance are caused by inevitable visual ambiguities in terms of appearance similarity among different people and visual variations of the same person under viewpoints, illumination, and occlusion.

The problem of re-identification has been extensively studied for still images by matching spatial appearance features (\eg color and texture) in correspondence using a pair of static images \cite{Farenzena2010Person,MidLevelFilter,Gray2008Viewpoint,eSDC,Pedagadi2013Local,LOMOMetric,Con-adaptation-cviu}. However, less attention has been paid to video-based approaches. In practice, video-based person re-id provides a more natural solution to person recognition because videos of pedestrians can be easily captured in a surveillance system. Furthermore, videos contain richer information than a single image, and beneficial for identifying a person under complex conditions. Given a bunch of frames in sequences, temporal priors in relation to some person motion can be captured and may assist in disambiguating an impostor across camera views. Also, sequential frames provide more samples for pedestrian appearance, where each sample may contain different poses and viewpoints and allow a reliable appearance based model to be constructed. However, making use of videos brings new challenges to re-id, namely the demand of dealing with time series with variable length and/or different frame rates. Moreover, the problem of seeking discriminative features from the videos with partial or full occlusion \cite{RCNRe-id} also needs to be tackled.

\begin{figure}[t]
\centering
\begin{tabular}{c}
\includegraphics[width=7.5cm,height=4cm]{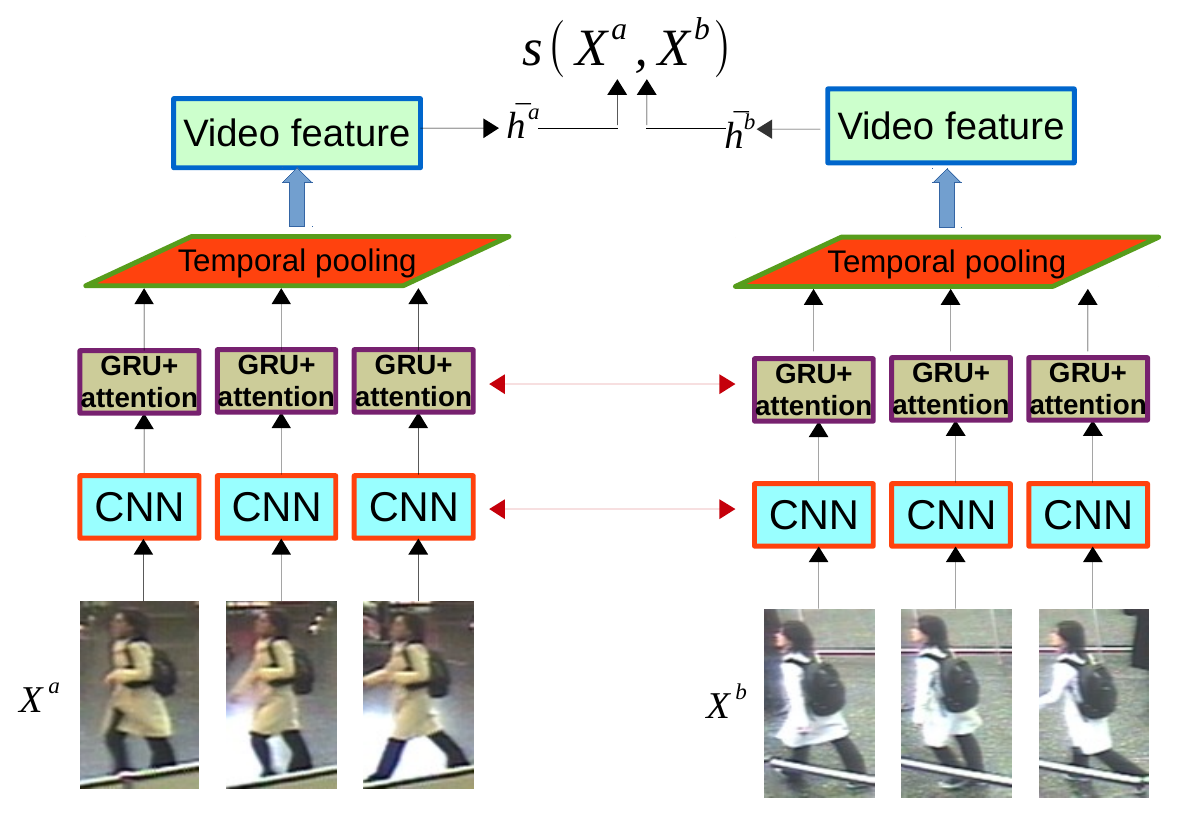}
\end{tabular}
\caption{Given a pair of pedestrian videos with variable length ($X^a$ and $X^b$), the proposed deep Siamese attention model incorporates CNNs, gated recurrent units and temporal pooling over time to jointly learn spatiotemporal video features and their corresponding similarity value ($s(X^a, X^b)$). Horizontal red arrows indicate that two subnetworks share the same parameterizations.}\label{fig:framework}
\end{figure}

In general, video based person re-id methods involve two key steps: feature learning and metric learning. Feature learning algorithms describe persons in videos by ascribing to spatial and temporal cues. The spatial part carries the information about visual appearance in correspondence while the temporal term is complementary to spatial alignment by providing motion across frames. To explore spatiotemporal information, a number of approaches \cite{VideoRanking,VideoPerson,Video-person-matching,RFA-net} are developed to extract low-level features from manually aligned sequences. They commonly extract frame-wise low-level 3D features (\eg HOG3D \cite{VideoRanking,HOG3D}, color/gradient features over color channels \cite{VideoPerson}) through pre-aligned sequences and aggregate these features via an encoding strategy. While metric learning refers to develop metrics through which the similarity between two matched videos of the same pedestrian is higher than that between videos of different pedestrians \cite{Video-person-ijcai16,Top-push}. Recent advances in Recurrent Neural Networks (RNNs) and the variants of Long Short-Term Memory (LSTM) models \cite{VideoLSTM} provide the insights on how to leverage temporal and spatial information for learning video representations. Inspired by the effectiveness of RNNs in sequence modeling, video-based person re-id gains some progress by using LSTM networks to learn feature representations \cite{RCNRe-id,RFA-net} or learn metrics \cite{Video-end-end}. However, these methods still suffer from certain limitations. First, the recurrence is only applied at the temporal dimension across frames while spatial correlation within each frame is merely revealed by performing convolutional operations. As a result, local features are independently extracted from each region whereas spatially contextual connections between regions are not captured, which are very crucial to matching persons \cite{Zhao2013SalMatch,S-LSTM,Wu-cyber18}. Second, the hidden LSTM unit is defined as a fixed-dimensional vector, which does not have the flexibility in choosing relevant contexts in spatial correlations when measuring the similarity between video pairs. Last but not the least, those RNN methods pay attention to learn video features or similarity metrics independently.

\subsection{Our Approach}
In this paper, we introduce a novel deep attention based Siamese model to video-based person re-id by jointly learning spatiotemporal video representations and similarity metrics. The network is designed to improve the discriminative capability of local features by leveraging the spatially contextual features within frames into the representation learning. In this sense, the similarity between two corresponding locations in a pair of videos is the integration of spatial context. We achieve the spatial recurrence by proposing convolutional gated recurrent units (GRUs) \cite{GRU2014} to propagate their relevance across temporal dimensions. The self-recurrent connections in the GRUs enable them to learn representations from the inputs that have been ``seen''. Moreover, the hidden units in our network are modified to be 3D feature cubes, which are flexible in capturing the varied input signals.

However, not all the spatial dependencies are relevant for the pedestrian matching, and indiscriminately scanning all regions incurs very high computational cost. To this end, we introduce the attention mechanism between the local convolutional activations and hidden states to enable the propagation of certain spatial contexts with discriminations and block the irrelevant ones. As a result, video fragments are not compressed into static hidden representations, but dynamically pooled out to enhance the discriminative capability of local features. Generally, for the metric learning of person re-id, feature vectors of similar pairs from the same subject must be ``close'' to each other while those from dissimilar pairs should be distant. To jointly learn similarity metrics, the proposed Siamese networks consist of two identical subnetworks which are integrated at the last convolution operations for comparing the input videos \cite{Siamese}. To optimize the set of network parameters, inputs are given in the form of pairs, and our network can be trained to minimize the cross-entropy loss on the input pair.

Formally, the Siamese architecture takes a pair of videos $X^a = \{x_1^a, x_2^a, \dots, x_{T_a}^a\}$, $X^b = \{x_1^b, x_2^b, \dots, x_{T_b}^b\}$ in their RGB values along with a class label $y^a$ ($y^b$) for each video. At each time-step, frames $x_t^a$ ($x_t^b$) are processed by Convolutional Neural Networks (CNNs) to produce their convolutional features. These local features are then weighted by a soft attention to capture their spatial correlations via a probability distribution on spatial dimensions. The resulting features are fed into convolutional GRUs to propagate the most relevant features over time. This soft attention is deterministic to enable efficient training using back-propagation, and potentially determines which local regions should be focused with higher importance being attached. Finally, an average temporal pooling is applied on hidden states across all time steps to generate the video-level representations $\bar{h}^a$ ($\bar{h}^b$). The similarity between $X^a$ and $X^b$, \ie $s(X^a,X^b)$, is defined as a weighted inner product between their video representations. The network can be trained by minimizing the cross-entropy loss on pairs of similar ($y^a=y^b$) and dissimilar time series ($y^a\neq y^b$). The overall architecture is shown in Fig. \ref{fig:framework}.

\subsection{Contributions}
The major contributions of this paper are summarized below.
\begin{itemize}
\item We propose a deep Siamese attention architecture to jointly learn feature representations and similarity metrics for video-based person re-id. The network leverages spatial contexts to enhance discriminative capabilities of local features and propagate critical spatial dependencies in computing the similarity of video pairs.
\item We introduce attention mechanism to select the most relevant features during the recurrence, which learns to attend at distinct regions in cross-view and help re-identify persons under complex conditions. Our method discloses which elements are important in video-based matching by visualizing ``where'' and ``when'' the attention focused on, and how to dynamically pool out the relevant features.
\item We conduct extensive experiments to demonstrate the state-of-the-art performance achieved by our method for video-based person re-id.
\end{itemize}

The rest of this paper is structured as follows. In Section \ref{sec:related}, we briefly review related works in terms of video-based person re-id, deep learning and visual attention. The proposed method is presented in Section \ref{sec:attention}, followed by parameter learning and experimental evaluations in Section \ref{sec:training} and \ref{sec:exp}, respectively. Section \ref{sec:con} concludes this paper.

\section{Related Work}\label{sec:related}

\subsection{Person Re-identification}
Many approaches to person re-id are image-based, which can be generally categorized into three categories. The first pipeline of these methods is appearance based design that aims to extract features that are discriminative and invariant against dramatic visual changes across views \cite{Farenzena2010Person,MidLevelFilter,eSDC}. The second stream is based on metric learning which works by extracting features for each image first, and then learning a metric with which the training data have strong inter-class differences and intra-class similarities \cite{LADF,LOMOMetric,KISSME}. Deep neural network approaches to person re-id jointly learn reliable features and corresponding similarity value for a pair or triplet of images \cite{PersonNet,DeepRanking,CNN-Re-id-TOMM,Transfer-re-id}. The idea is to directly comparing pairs of images and answer the question of whether the two images depict the same person or not. Another pipeline of methods in person re-id are post-ranking based optimization \cite{POP,Post-ranking} which can exploit the context information from the first ranked positions to detect/remove visual ambiguities to compute an improved new ranking. However, these approaches are developed to work on still image based person re-id, whilst not applicable to the video setting.

Many realistic scenarios indicate that multiple images can be exploited to improve the matching performance. Multi-shot approaches in person
re-id \cite{Farenzena2010Person,VideoRanking,PaMM} use multiple images of a person to extract appearance descriptors to model person appearance. For these methods, multiple images from a sequence are used to enhance spatial feature descriptions from local regions \cite{Farenzena2010Person}. Some methods attempt to select and match video fragments to maximize the cross-view ranking \cite{VideoRanking}. These methods, however, deal with multiple images independently whereas in the video context, a video contains more information than independent images, \eg dynamics of motion and temporal evolution. Learning temporal dynamics in person re-id with videos is very challenging, and earlier approaches use optical flow, HOG and hand-crafted features to generate descriptors with both appearance and dynamics information encoded. These spatiotemporal appearance models \cite{Gheissari2006Person,VideoPerson} treat the video data as a 3D volume and extract local spatiotemporal features. They construct spatiotemporal volume based representations by extending image descriptors such as 3D-SIFT \cite{3DSIFT} and HOG3D \cite{HOG3D}. In \cite{VideoRanking}, HOG3D is utilized as a spatiotemporal feature representation for person re-id due to its robustness against cluttered backgrounds and occlusions. Liu \etal \cite{VideoPerson} proposed a better representation that encodes both the spatial layout of the body parts and the temporal ordering of the action primitives of a walking pedestrian. However, these methods construct representations from low-level descriptors including color, texture and gradients, which are still not discriminate enough to distinguish persons and very sensitive to large intra-person visual variations. Also, the temporal nature of videos is not explicitly modeled in the their approaches.  A recent approach proposed by McLaughlin \etal \cite{RCNRe-id} combines CNNs with LSTMs and performs temporal pooling of features prior to classifying sequences into proper identities. A similar work is presented in \cite{RFA-net} where low-level features are fused under LSTMs.

In this paper, we deliver an attention based Siamese model to jointly learn spatiotemporal expressive features and corresponding similarity metric for a pair of pedestrian sequences. Our architecture improves the discriminative capability of local features by encoding spatial contexts via attention selection. This also regulates the propagation of the most relevant contexts through the network.

\subsection{Deep Learning and Visual Attention}

Deep learning methods in person re-id are effective in combating visual variations by learning a projection into deep embedding space under which the identity information can be preserved \cite{GAN-re-id,SVDNet,CNN-Re-id-TOMM}. And the learned representations can also be transferred to different datasets \cite{Re-id-ICDSC,Transfer-re-id}. However, these deep models are still difficult to be examined to understand the learned capability of patterns in person recognition. The principled saliency based methods \cite{eSDC,KEPLER} resemble human gazing capabilities to identify the salient regions of a person appearance to tackle this challenging problem. We are inspired by saliency methods in detecting informative regions whilst we attempt the first efforts in localizing regions both spatially and temporally over a video sequence.

Recent deep attention models add a dimension of interpretation by capturing where the model is focusing its attention when performing a particular task \cite{Attention-Encoder}. For instance, Xu \etal \cite{ShowAttendTell} used both soft attention and hard attention mechanisms to generate image descriptions. Their models look at the perspective objects when generating their descriptions. While Xu \etal \cite{ShowAttendTell} primarily work on caption generation in static images, in this paper, we focus on using a soft attention mechanism for person re-id in videos. A similar work is from \cite{ActionAttention} where an attention model is presented to classify actions in videos. However, pedestrians in the wild exhibit similar walking behaviors without distinctive and semantically categorisable actions unique to different people.

\begin{figure}[t]
\begin{tabular}{cc}
\includegraphics[width=4cm]{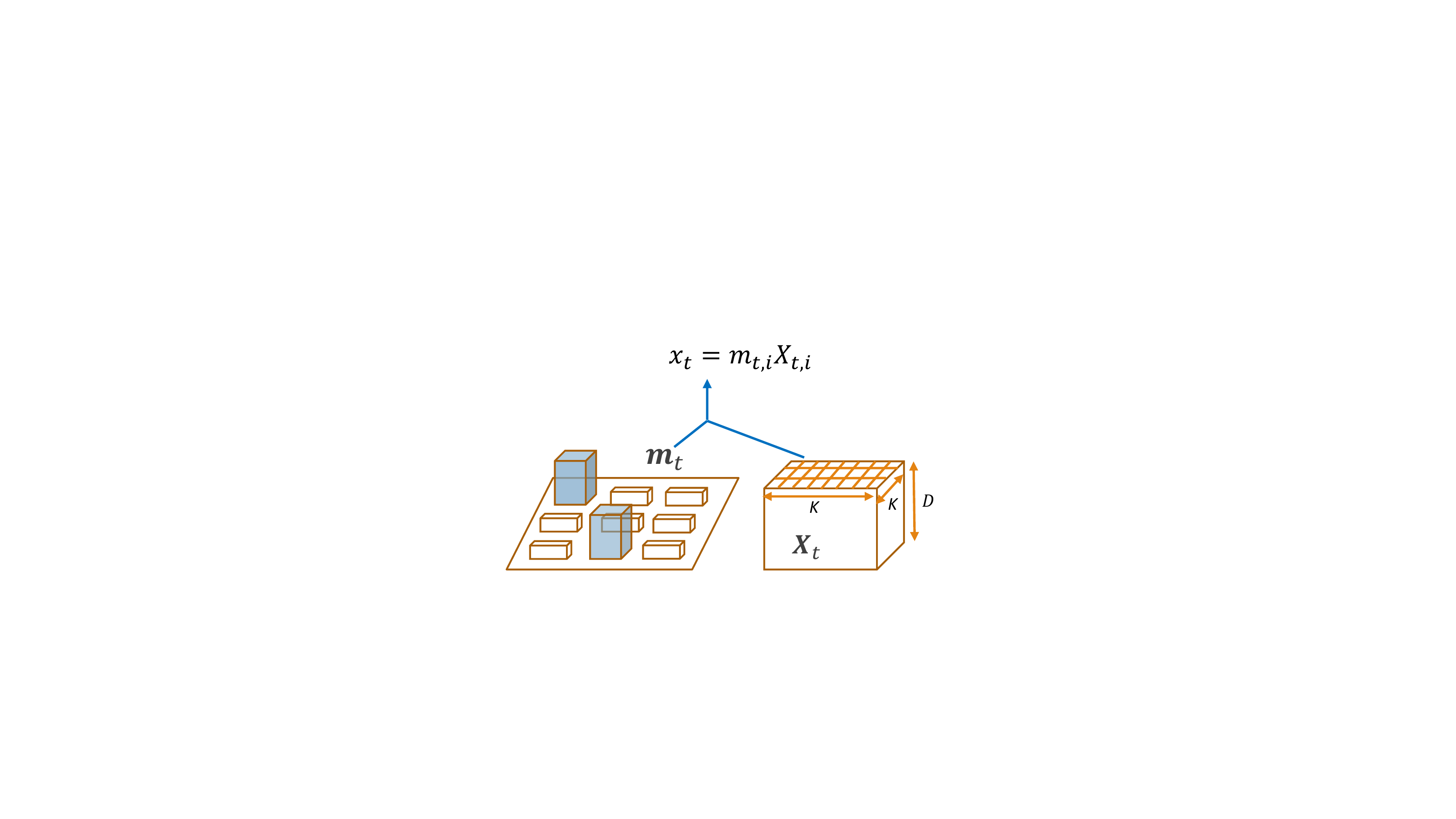}&
\includegraphics[width=4cm]{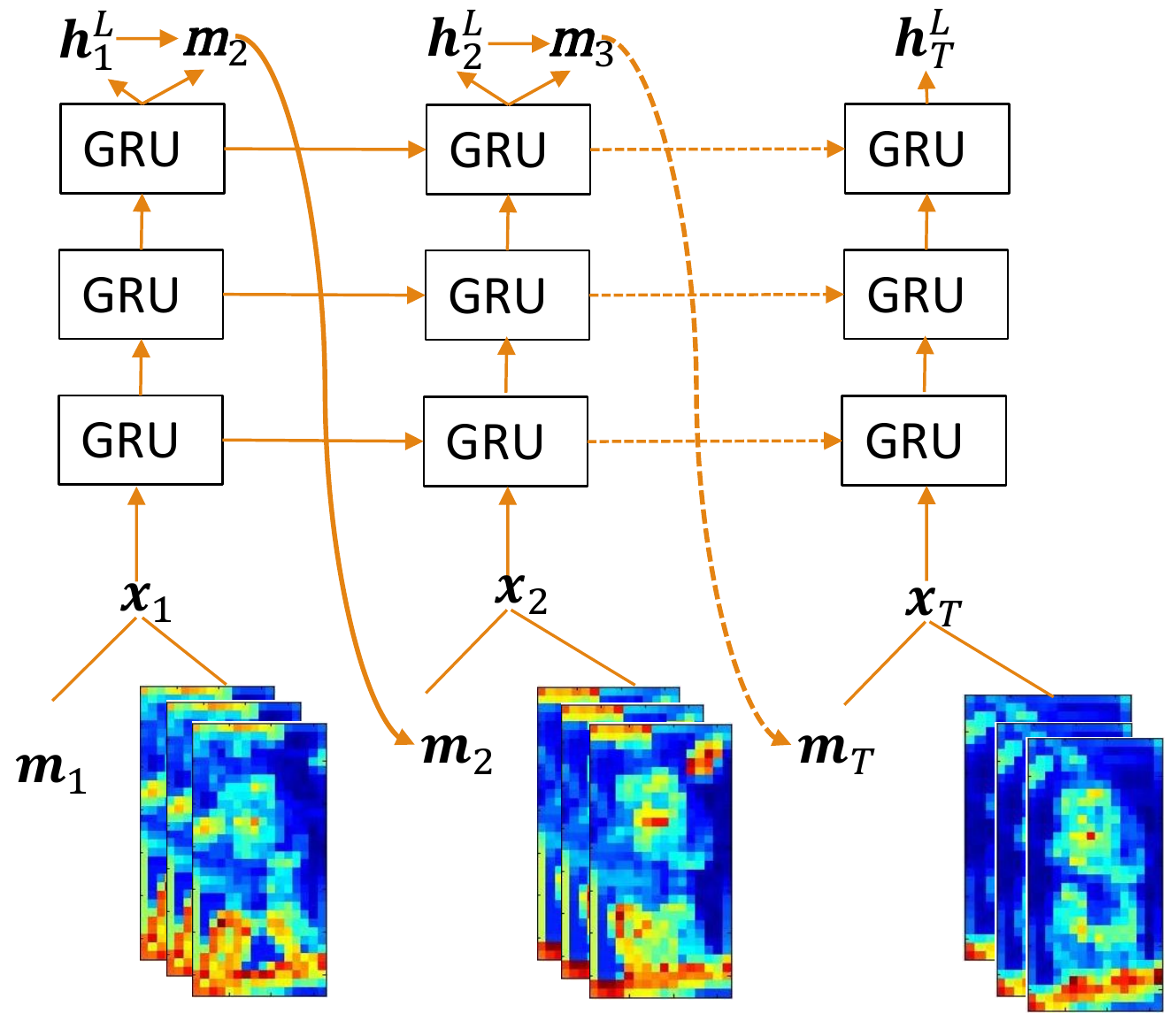}\\
(a)  & (b)
\end{tabular}
\caption{(a) The CNN encoder takes the video frame as its input, and produces a feature cube $\bX_t\in \mathbb{R}^{K\times K \times D}$, which is averaged and weighted on the location softmax $\bm_t$ to compute the input $\bx_t$ for the subsequent recurrent layer. (b) At each time step $t$, the feature cube $\bx_t$ is fed into three layers of GRU to propagate and predict the next location probabilities $\bm_{t+1}$ and hidden-representations $\bh_t$.}\label{fig:model}
\end{figure}

\section{Deep Siamese Attention Networks for Video-based Person Re-identification}\label{sec:attention}

In this section, we introduce attention based deep Siamese networks for video-based person re-id. The goal is to match sequences of same pedestrians obtained from different surveillance systems. The proposed Siamese architecture employs two convolution-recurrent networks that share their parameters in order to extract comparable hidden-unit representations from which a similarity value is computed as output. The network is optimized based on the cross-entropy loss. Fig. \ref{fig:framework} illustrates the proposed Siamese architecture. At each time step $t$, frames $x_t^a$ and $x_t^b$ of sequence $X^a$ and $X^b$ passes through the CNNs, the recurrent layers with spatial context encoding, and the temporal pooling layer, to jointly learn spatiotemporal video features and their similarity value $s(X^a,X^b)$. A notation table is given in Table \ref{tab:notation}.

\begin{table}[t]
\caption{Notations and definitions.}\label{tab:notation}
\centering
\scriptsize
\begin{tabular}{|c|c|}
\hline
Notation &  Definition \\
\hline
$X^a$, $X^b$ & Two sequences across view\\
$T$ & Total length of a sequence \\
$K\times K\times D$ & Convolutional feature cubes \\
$W$ & Input-to-hidden parameter\\
$U$ & Hidden-to-hidden parameter\\
$k_1 \times k_2$ & The convolutional kernel size\\
$\bz$, $\br$ & Update and reset gates\\
$\bh_t^l$ & Hidden representations at time-step $t$ of layer-$l$\\
$\bar{\bh}$ & The final pooled hidden video represenations\\
$m_{t,i}$ & Attention weight on location $i$ at time-step $t$\\
$\lambda$ & The attention penalty coefficient\\
\hline
\end{tabular}
\end{table}

\subsection{Convolutional Layers}

Our architecture employs Convolutional Neural Networks (CNNs) to transfer the input data to a representation where the learning of pedestrian motion dynamics is easy \cite{RNNHuman}. In general, CNNs process an image using a series of layers, where each individual layer is composed of convolution, pooling, and non-linear activation function steps. In our approach, we extract the last convolutional layer by passing the video sequence through GoogLeNet model \cite{GoogLeNet} trained on the ImageNet dataset \cite{ImageNet}. Let $X=[x_1,\ldots,x_T]$ be a video sequence of length $T$, consisting of whole-body images of a person, where $x_t$ is the image at time $t$. The last convolutional layer in GoogLeNet has $D$ convolutional maps and is feature cube of shape $K\times K\times D$ ($7\times7\times1024$ in our experiments). Thus, at each time-step $t$, we extract $K^2$ $D$-dimensional vectors. We refer to these vectors as feature slices in a feature cube:
\begin{equation}\small
\bX_t=[\bX_{t,1},\ldots,\bX_{t,K^2}], \bX_{t,i}\in \mathbb{R}^D.
\end{equation}
Each of these $K^2$ vertical slices links to different overlapping regions, and corresponds to portions of local regions in the 2-D frame. Our model tends to selectively focus on certain parts of these $K^2$ regions to propagate through the network.

\subsection{Deterministic Soft Attention Mechanism}\label{ssec:soft}
We use a particular type of RNN model, Gated Recurrent Units (GRUs) \cite{GRU2014} to learn an effective representation of person videos by capturing its long range dependencies. We implement GRUs as the recurrent units due to its superior performance in many tasks such as music/speech signal modeling \cite{EmpiricalGRU} and fewer parameters as opposed to the Long Short Term Memory (LSTM) \cite{lstm1997}.
The GRU is formulated as follows:
\begin{equation}\label{eq:fully_GRU}\scriptsize
\begin{split}
& \bz_t = \sigma(W_z \bx_t + U_z \bh_{t-1}), \br_t = \sigma(W_r \bx_t + U_r \bh_{t-1}),\\
& \hat{\bh}_t = tanh(W \bx_t + U(\br_t \odot \bh_{t-1})),\\
& \bh_t = (1 - \bz_t)\bh_{t-1} + \bz_t \hat{\bh}_t,
\end{split}
\end{equation}
where $\odot$ is an element-wise multiplication. $\bz_t$ is an update gate that determines the degree to which the unit updates its activation. $\br_t$ is a reset gate, and $\sigma$ is the sigmoid function. The candidate activation $\hat{\bh}_t$ is computed similarly to that of traditional recurrent unit. When $\br_t$ is close to zero, the reset gate make the unit act as reading the first symbol of an input sequence and forgets the previously computed state.

The vector $\bx_t$ is a dynamic representation of the relevant part of the frame input at time $t$. To compute $\bx_t$ from the feature cube $\bX_{t,i}, i=1,\ldots, K^2$ corresponding to features extracted at different image regions, our model defines a mechanism to predict $\bm_{t+1}$, a softmax over $K\times K$ locations. For each location $i$, the mechanism generates a positive weight $m_{t,i}$ which can be interpreted as the relative importance to give to location $i$ in blending the $\bX_{t,i}$'s together ({\it deterministic attention mechanism}). The weight $m_{t,i}$ of each feature slice $\bX_{t,i}$, \ie location softmax, is computed by an attention model for which we use a multi-layer perception conditioned on the previous hidden state $\bh_{t-1}$. The location softmax is defined as follows:
\begin{equation}\label{eq:location_softmax}\small
m_{t,i}=p(\bL_t=i|\bh_{t-1})=\frac{\exp (W_i^T \bh_{t-1})}{\sum_{j=1}^{K\times K} (W_j^T \bh_{t-1})}, i\in 1, \ldots, K^2,
\end{equation}
where $W_i$ are weights mapping to the $i$-th element of the location softmax and $\bM_i$ is a random variable which can take 1 of $K^2$ values. This softmax can be interpreted as the probability from which our model suggests the corresponding region in the input frame is important \cite{ActionAttention}. We emphasize that the hidden states varies as the RNN output advances in its output sequence: ``where'' the network looks next depends on the sequence of regions that has been already examined.

Within the probabilities of locations, the deterministic soft attention mechanism computes the expected value of the input at the next time-step $\bx_t$ by taking weighted mask over the feature slices at different regions. A graphical depiction of deterministic soft attention mechanism is shown in Fig.\ref{fig:model} (a). Hence, the deterministic attention model by computing a soft attention weighted feature cube is formulated as
\begin{equation}\label{eq:soft_attention}
\bx_t= m_{t,i} \bX_{t,i}.
\end{equation}

This corresponds to feeding a soft weighted context into the system. Note that an alternative mechanism for attention is stochastic ``hard'' attention where the location variable $\bL_t$ should be sampled from the softmax distribution of Eq.\eqref{eq:location_softmax} \cite{ShowAttendTell}. However, hard attention models are not differentiable and have to resort to some form of sampling. By contrast, a soft attention model as a whole is smooth and differentiable under the deterministic attention, so that learning end-to-end is trivial by using standard back-propagation.

\subsection{Deep Attention based Spatial Correlation Encoding}\label{ssec:deep_rcn}

Given the attention weighted feature cube $\bx_t$, we apply a stack of $L$-layer RNNs on each $\bx_t$ to propagate their attention over time. The stacked RNNs can be parameterized as $\phi^1$, $\phi^2$, $\dots$, $\phi^L$, such that $\bh_t^l=\phi^l(\bx_t, \bh_{t-1}^l), l=1,\ldots,L$. The output feature representations $\bar{\bh}$ are obtained by average/attention pooling the hidden unit activations over all time steps $\bh_1^L,\bh_2^L,\dots, \bh_T^L$, as defined in Eq.\eqref{eq:temporal-selection}. The recurrent function $\phi^l$ is implemented by using Gated Recurrent Units as formulated in Eq.\eqref{eq:fully_GRU}. However, the formulation of GRU is a \textit{fully-connected} GRU (FC-GRU) where the input and states are all 1-D vectors. To model the spatiotemporal relationship in video sequence in the form of convolutional activations, we restructure the fully-connected GRUs to be convolutional (ConvGRUs) which perform convolutional operations in both the input-to-hidden and hidden-to-hidden transitions.

\begin{figure}[t]
\centering
\includegraphics[height=3cm]{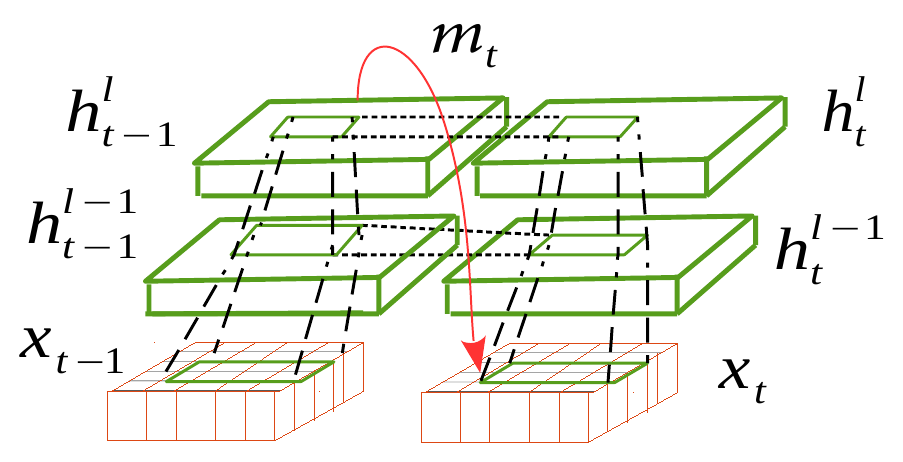}
\caption{The structure of a convolutional GRU. $\bx_t= m_{t,i} \bX_{t,i}$ have dimension $K\times K\times D$, corresponding to the frame-level CNN features with attention applied. To encode the spatial information into hidden state $\bh_t^l$, the state in the grid is determined by the input ($\bx_t$) and the past state of its local neighbors ($\bh_{t-1}^l$), as implemented by a convolutional operation. }
\label{fig:ConvGRU}
\end{figure}

The output convolutional maps of the encoder subordinate to recurrent layers are 3D tensors (spatial dimensions and input channels) for which directly applying a FC-GRU can lead to a massive number of parameters and too much redundancy for these spatial data. Let $H_1$ and $H_2$ and $C_x$ be the input convolutional map spatial dimension and number of channels. A FC-GRU would require the input-to-hidden parameters $W_z^l$, $W_r^l$ and $W^l$ of size $H_1 \times H_2 \times C_x \times C_h$ where $C_h$ is the dimensionality of the GRU hidden representation. Inspired by Convolutional LSTMs \cite{ConvLSTM}, we employ convolutional GRUs to capture the motion dynamics  involved over time series. The ConvGRUs are first introduced by Ballas \etal \cite{ConvGRU} to take advantage of the underlying structure of convolutional maps.
The rationale of using ConvGRUs is based on two observations: 1) convolutional maps extracted from images represent strong \textbf{local correlations} among responses, and these local correlations may repeat over different spatial locations due to the shared weights of convolutional filters (a.k.a. kernels) across the image spatial domain; 2) videos are found to have strong \textbf{motion smoothness} on temporal variation over time, \eg motion associated with a particular patch in successive frames will appear and be restricted within a local spatial neighborhood. In this sense, we reduce the computational complexity of the deep RNNs by further sparsifying their hidden unit connection structures. Specifically, our model is designed to characterize the feature locality and motion smoothness, and the fully-connected units in GRUs are replaced with convolutional filters (kernels). Thus, recurrent units are encouraged to have sparse connectivity (hidden-to-hidden) and share their parameters across different input spatial locations (input-to-hidden), and we can formulate a ConvGRU as:
\begin{equation}\label{eq:ConvGRU}\scriptsize
\begin{split}
& \bz_t^l = \sigma(W_z^l * \bh_t^{l-1} + U_z^l * \bh_{t-1}^l), \br_t^l = \sigma(W_r^l * \bh_t^{l-1} + U_r^l * \bh_{t-1}^l),\\
& \hat{\bh}_t^l = tanh(W^l * \bh_t^{l-1} + U^l * (\br_t^l \odot \bh_{t-1}^l )),\\
& \bh_t^l = (1- \bz_t^l)\bh_{t-1}^l + \bz_t^l \hat{\bh}_t^l; l=L,\ldots,1,
\end{split}
\end{equation}
where * denotes a convolution operation and $\bh_t^0 = \bx_t$.  In Eq \eqref{eq:ConvGRU}, input-to-hidden parameters $W_z^l$, $W_r^l$, $W^l$, and hidden-to-hidden parameters $U_z^l$, $U_r^l$, $U^l$ are 2-D convolutional kernels with size of $k_1\times k_2$ where $k_1\times k_2$ is much smaller than the original convolutional map size $H_1 \times H_2$. The input and hidden representation can be imagined as vectors standing on spatial grid, and the ConvGRU determines the future hidden representation in the grid by the inputs and past hidden units of its local neighbors. The resulting hidden recurrent representation $\bh_t^l = \{\bh_t^l(i,j)\}$ can preserve the spatial structure where $\bh_t^l(i,j)$ is a feature vector defined at the grid location $(i,j)$. To ensure the spatial size of the hidden representation remains the same over time, zero-padding is needed before applying the convolution operation. The structure of a convolutional GRU is illustrated in Fig.\ref{fig:ConvGRU}. In fact, the hidden recurrent representation of moving objects is able to capture faster motion moving with a larger transitional kernel size while one with a smaller kernel can capture slower motion. Recall that FC-GRU models inputs and hidden units as vectors, which can be viewed as a feature map with $1\times 1$ spatial dimension. In this sense, FC-GRU becomes a special case of ConvGRU with all features being a single vector. The advantage of adding the deep hidden connections is to enable the model to leverage representations with different resolutions since these low-level spatial resolutions certainly encode informative motion patterns. In particular, the receptive field associated with each $\bh_t^l (i,j)$  increases in previous hidden states $\bh^l_{t-1}$, $\bh^l_{t-2}$,\dots, $\bh^l_{1}$ if we go back along the time line.

\subsection{Temporal Pooling: Attention on Temporal Selection}

A temporal pooling is needed in our architecture because RNN's outputs may be biased towards later time-steps, which could reduce the effectiveness of RNN when used to summarize the relevant information over a full sequence since discriminative regions can appear anywhere in the sequence. To address this limitation, an effective average temporal pooling is introduced on the top of the network to combine the features at all convolutional levels but from selected frames to generate overall feature appearance for the complete sequence. Specifically, we propose a soft-attention based temporal pooling, which is formulated as
\begin{equation}\label{eq:temporal-selection}\small
  \bar{\bh} = \frac{\alpha_t \bh_t^L}{\sum_{t=1}^T \alpha_t \bh_t^L},
\end{equation}
where $\boldsymbol{\alpha}=[\alpha_1, \cdots, \alpha_T], (\sum_{t=1}^T \|\alpha_t\|=1)$ is the selection vector that is able to re-weigh each frame to selectively focus on certain subset of frames. For each frame at time step $t$, the mechanism generates a positive weight $\alpha_t$, which can be interpreted as the relative importance to give to the frame at time $t$ in blending the $\bh_t^L$'s together. This corresponds to feeding a soft $\boldsymbol{\alpha}$ weighted temporal context into the system. And the whole model is smooth and differentiable under this soft deterministic attention, so learning end-to-end is still trivial by using standard back-propagation. Then, $\bar{\bh} $ is flattened to produce a single vector as its video-level representation. This allows for the aggregation of information across all time steps, thus avoiding bias towards later time-steps.

\subsection{Complexity}

Due to the convolution operation, input-to-hidden parameters $W_z^l$, $W_r^l$, $W^l$ have the size of $k_1\times k_2 \times C_x \times C_h$. The activation gate $\bz^l_t(i,j)$,  reset gate $\br^l_t(i,j)$, and candidate hidden representation $\hat{\bh}_t^l (i,j)$ are defined based on a local neighborhood of size $k_1\times k_2$ at the location $(i,j)$ on both the input data $\bx_t$ and the previous hidden units $\bh^l_{t-1}$. A ConvGRU layer conducts 2D convolutional operation 6 times at each time step \ie 2 convolutions per GRU gate and another 2 in computing $\hat{\bh}^l_t$. Assume the input-to-hidden and hidden-to-hidden transition parameters have the same kernel size while preserving the input dimensions, ConvGRU requires $O(3 T H_1 H_2 k_1 k_2 (C_x C_h + C_h^2))$ computations where $T$ is the length of a time sequence. Apparently, ConvGRU saves computations substantially compared with a FC-GRU which requires  $O(3 T H_1^2 H_2^2 (C_x C_h + C_h^2))$ multiplications. In memory, a ConvGRU only needs to store parameters for 6 convolutional kernels, yielding $O(3 k_1 k_2 (C_x C_h + C_h^2))$
parameters.

\section{Parameter Learning}\label{sec:training}

\subsection{Initialization}
We use the following initialization strategy \cite{ShowAttendTell} for the hidden states for faster convergence:
\begin{equation}\scriptsize
\bh_0= f_{init}\left( \frac{1}{T} \sum_{t=1}^T \left( \frac{1}{K^2} \sum_{i=1}^{K^2} \bX_{t,i}\right) \right),
\end{equation}
where $f_{init}$ is a multi-layer perception. These values are used to calculate the first softmax location $\bm_1$ which determines the initial input $\bx_1$.

\subsection{Loss Function and Attention Penalty}
We use cross-entropy loss coupled with the doubly stochastic penalty \cite{ShowAttendTell}. We impose an additional constraint over the location softmax, such that $\sum_{t=1}^T m_{t,i} \approx 1$. This is the attention regularization which forces the model to look at each region of the frame at some point in time. The cross-entropy loss can be computed from the similarity value between two time-series input.
Given two time series $X^{a}=\{x_1^a,x_2^a,\dots,x_{T_a}^a\}$ and $X^{b}=\{x_1^b,x_2^b,\dots,x_{T_b}^b\}$, the hidden unit representations $\bar{\bh}^a$ and $\bar{\bh}^b$ computed from the two subnetworks can be combined to compute the prediction for the similarity of the two time series. Thus, we define the similarity of the two sequences as:
\begin{equation}\label{eq:similarity}\small
s(X^a,X^b) = \frac{1}{1 + e^{-\bv^T [\mbox{diag}\left(\bar{\bh}^a ( \bar{\bh}^b)^{\mathsf{T}}\right)] + c}},
\end{equation}
where the element-wise inner product between the hidden representations is computed \ie $\bar{\bh}^a (\bar{\bh}^b)^{\mathsf{T}}$, and the output is a weighted sum of the resulting product ($\bv$ is the weight vector). $c$ is a constant value. The similarity between two time series is defined as a weighted inner product between the representations $\bar{\bh}^a$ and $\bar{\bh}^b$. Our approach learns a vectorial representation for each time series in such a way that similar/dissimilar time series are modeled by similar/dissimilar representations. Thus, time series modeling and metric learning can be combined and studied through the Siamese recurrent networks, which can be optimized to minimize the cross-entropy loss on pairs to learn a good similarity measure between time series.

Let $Z$ denote a training set containing two sets of pairs of time series: a set with similar pairs $Sim$ and a set with dissimilar pairs of $Dis$. We aim to learn all parameters $\Theta =\{ W_z^l, W_r^l, W^l, U_z^l, U_r^l, U^l, \bv, c\}$ in our network jointly by minimizing the binary cross-entropy prediction loss coupled with the penalty on location softmax. This is equivalent to maximizing the conditional log-likelihood of the training data:
{\scriptsize \begin{multline}\label{eq:loss}
   \mathcal{L}(\Theta; Z) = \lambda \sum_{i=1}^{K^2} (1- \sum_{t=1}^T m_{t,i})^2
    \\
    -\left[ \sum_{(a,b)\in Sim} \log s(X^a, X^b)
    + \sum_{(a,b)\in Dis}  \log (1-s(X^a, X^b))
       \right],
\end{multline}}
where $a$ and $b$ denote the index of two time sequences in training pairs, and $\lambda$ is the attention penalty coefficient.

The loss function is maximized and back-propagated through stacked recurrent networks (the weights of two subnetworks are shared) using a variant of the back-propagation through time algorithm with gradient clipping between -5 and 5 \cite{AdvanceRNN}. The sets $Sim$ and $Dis$ are defined using class labels for each time sequence: $Sim=\{ (a,b): y^a = y^b\}$ and $Dis=\{(a,b): y^a \neq y^b\}$ where $y^a$ and $y^b$ are class labels for time sequences $X^a$ and $X^b$. In the case of person re-id, each person can be regarded as a class, and thus training class label can be assigned on each sequence accordingly. In contrast to existing classification deep models in person re-identification \cite{CNN-Re-id-TOMM,PersonNet}, our loss function allows our architecture to be applied on objects from unknown classes. Our network can be trained one dataset (domain) and applied on a different test dataset (out-of-domain) to verify the new fragment sequences that were not present in training set. Thus, this loss function is more suitable to person re-id where the underlying assumption is that inter-person variations have been modeled well. The experimental analysis on cross-domain testing is given in Section \ref{ssec:cross-dataset}.

\section{Experiments}\label{sec:exp}

In this section, we present empirical evaluations on the proposed model. Experiments are conducted on three benchmark video sequences for person re-identification.

\subsection{Datasets}

\begin{figure}[t]
\includegraphics[height=2cm,width=3cm]{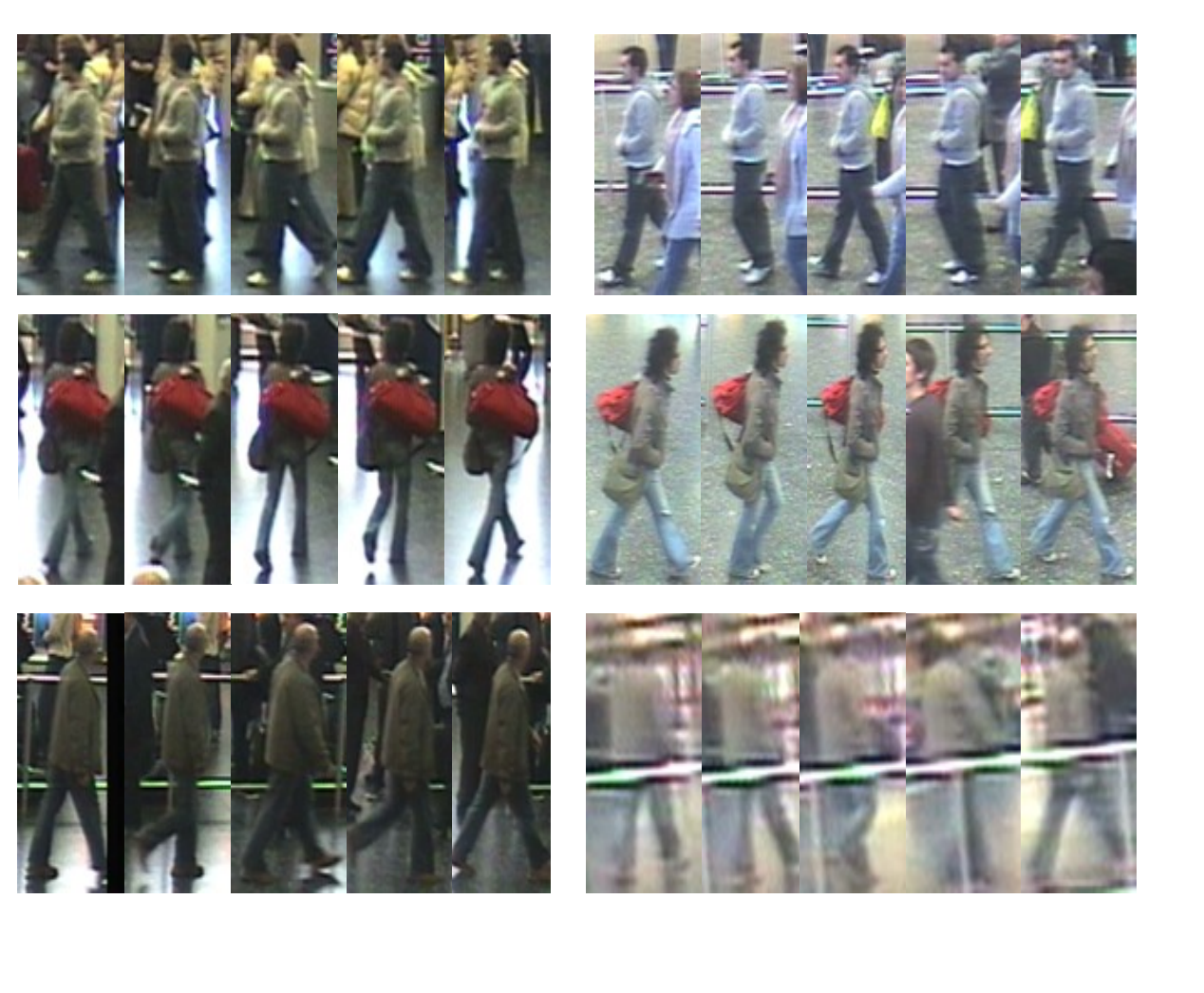}
\includegraphics[height=2cm,width=3cm]{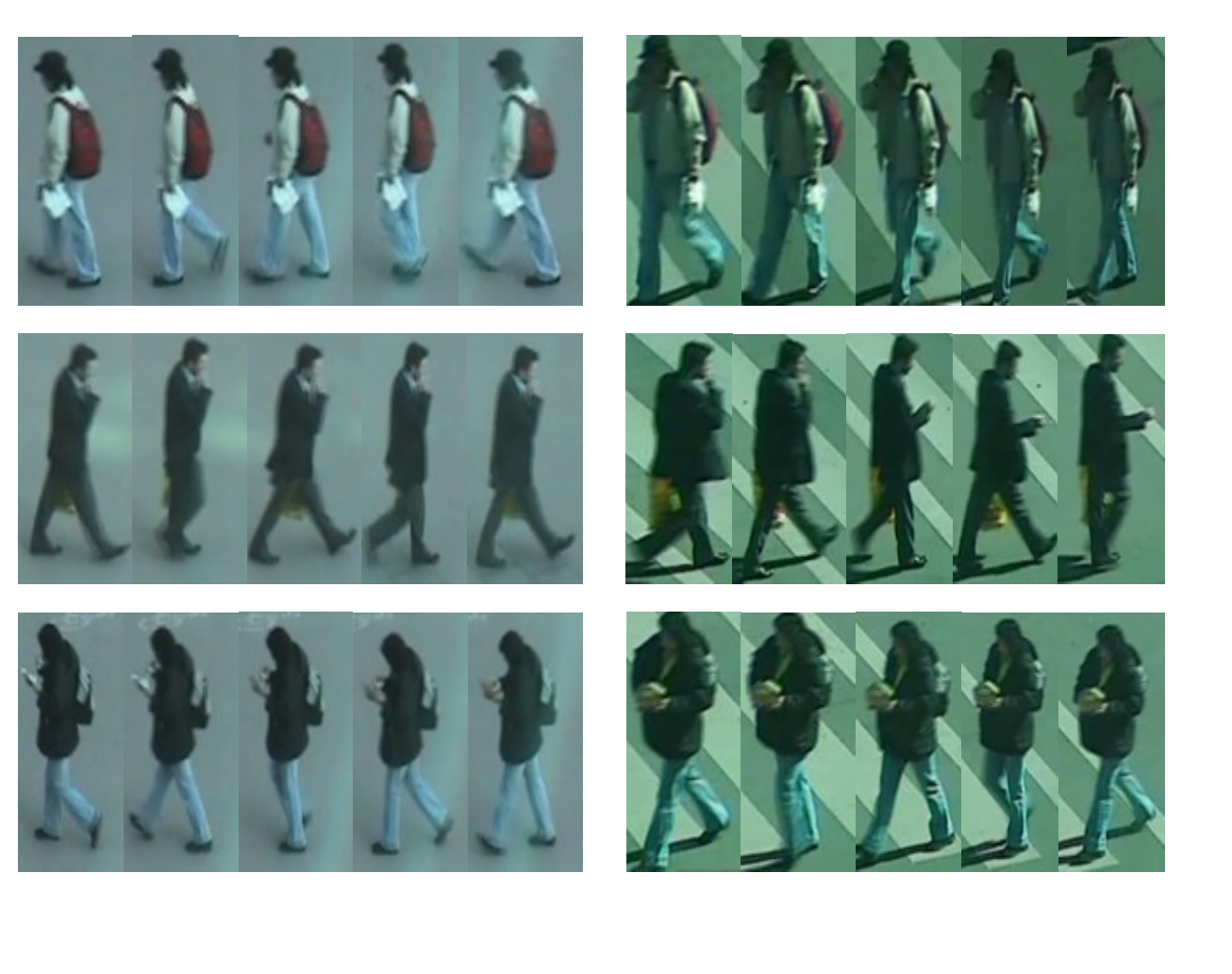}
\includegraphics[height=2cm,width=2.2cm]{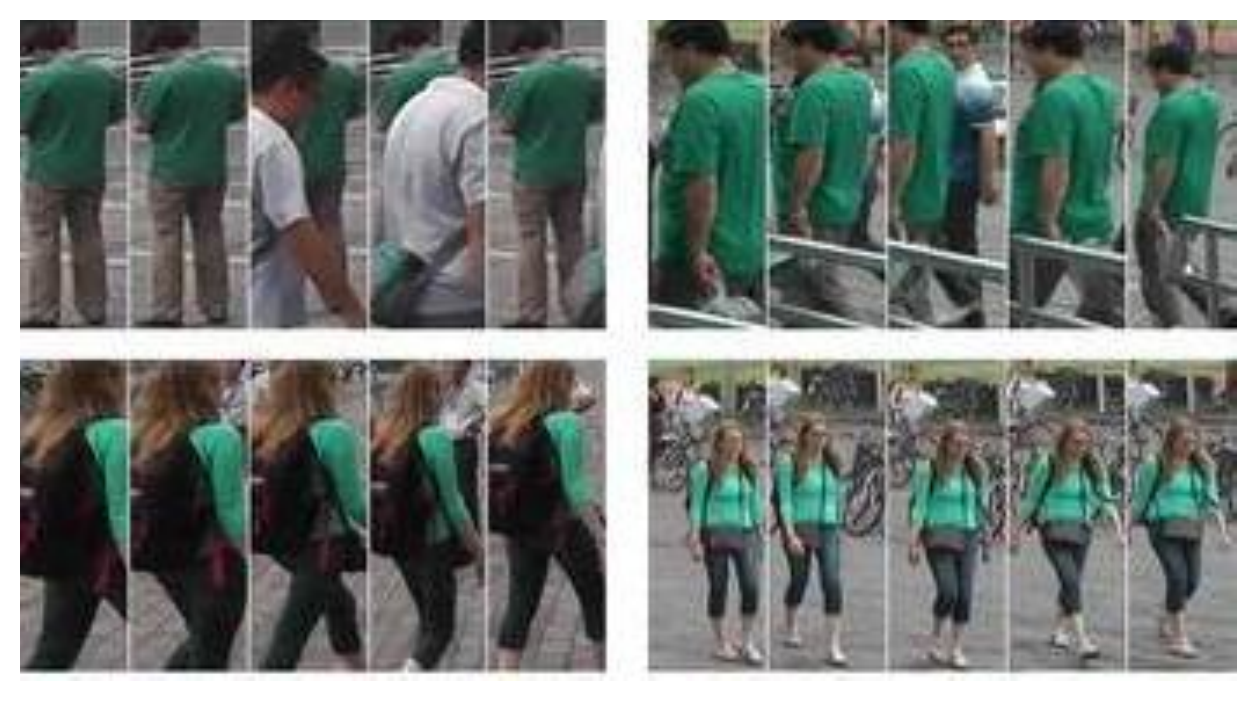}\\
(a) iLIDS-VID \hspace{1cm} (b) PRID2011 \hspace{1cm} (c) MARS
\caption{Image sequences of the same pedestrian (in row) in different camera views from the three datasets.}
\label{fig:examples}
\end{figure}

We experimented on three image sequence datasets (Fig.\ref{fig:examples}): iLIDS-VID \cite{VideoRanking}, PRID 2011 \cite{PRID2011}, and MARS \cite{MARS}.

\begin{itemize}
\item The iLIDS-VID dataset consists of 600 image sequences for 300 randomly sampled people, which was created based on two non-overlapping camera views from the i-LIDS multiple camera tracking scenario. The sequences are of varying length, ranging from 23 to 192 images, with an average of 73. This dataset is very challenging due to variations in lighting and viewpoint caused by cross-camera views, similar appearances among people, and cluttered backgrounds.
\item The PRID 2011 dataset includes 400 image sequences for 200 persons from two adjacent camera views. Each sequence is between 5 and 675 frames, with an average of 100. Compared with iLIDS-VID, this dataset was captured in uncrowded outdoor scenes with rare occlusions and clean background. However, the dataset has obvious color changes and shadows in one of the views.
\item The MARS dataset contains 1,261 pedestrians, and 20,000 video sequences, making it the largest video re-id dataset. Each sequence is automatically obtained by the Deformable Part Model \cite{DetectionPAMI} detector and the GMMCP \cite{GMMCP} tracker. These sequences are captured by six cameras at most and two cameras at least, from which each identity has 13.2 sequences on average. This dataset is evenly divided into train and test sets, containing 625 and 636 identities, respectively.
\end{itemize}

\subsection{Model Training}

We implemented our architecture in Python with Theano \cite{TheanoNew}. All experiments are run on a single PC with a single NVIDIA GTX 980 GPU with 12GB memory. For the iLIDS-VID and PRID 2011 datasets, the whole dataset is randomly split into 50\% (50\%) of persons for training (testing). For the MARS dataset, we comply with the fixed training and testing sets, containing 631 and 630 identities, respectively.

\begin{figure}[t]
\begin{tabular}{cc}
\includegraphics[height=3cm]{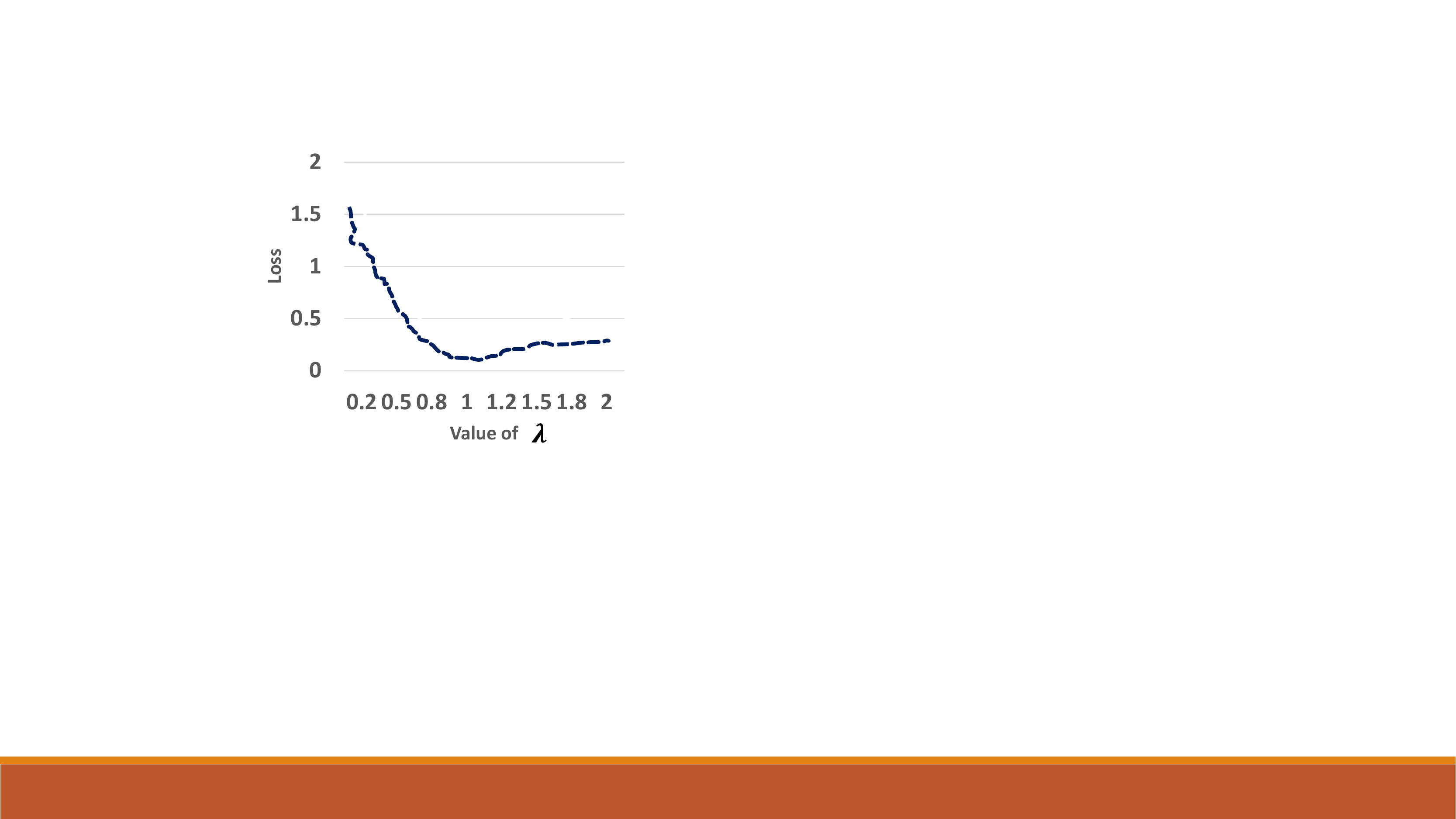}&
\includegraphics[height=3cm]{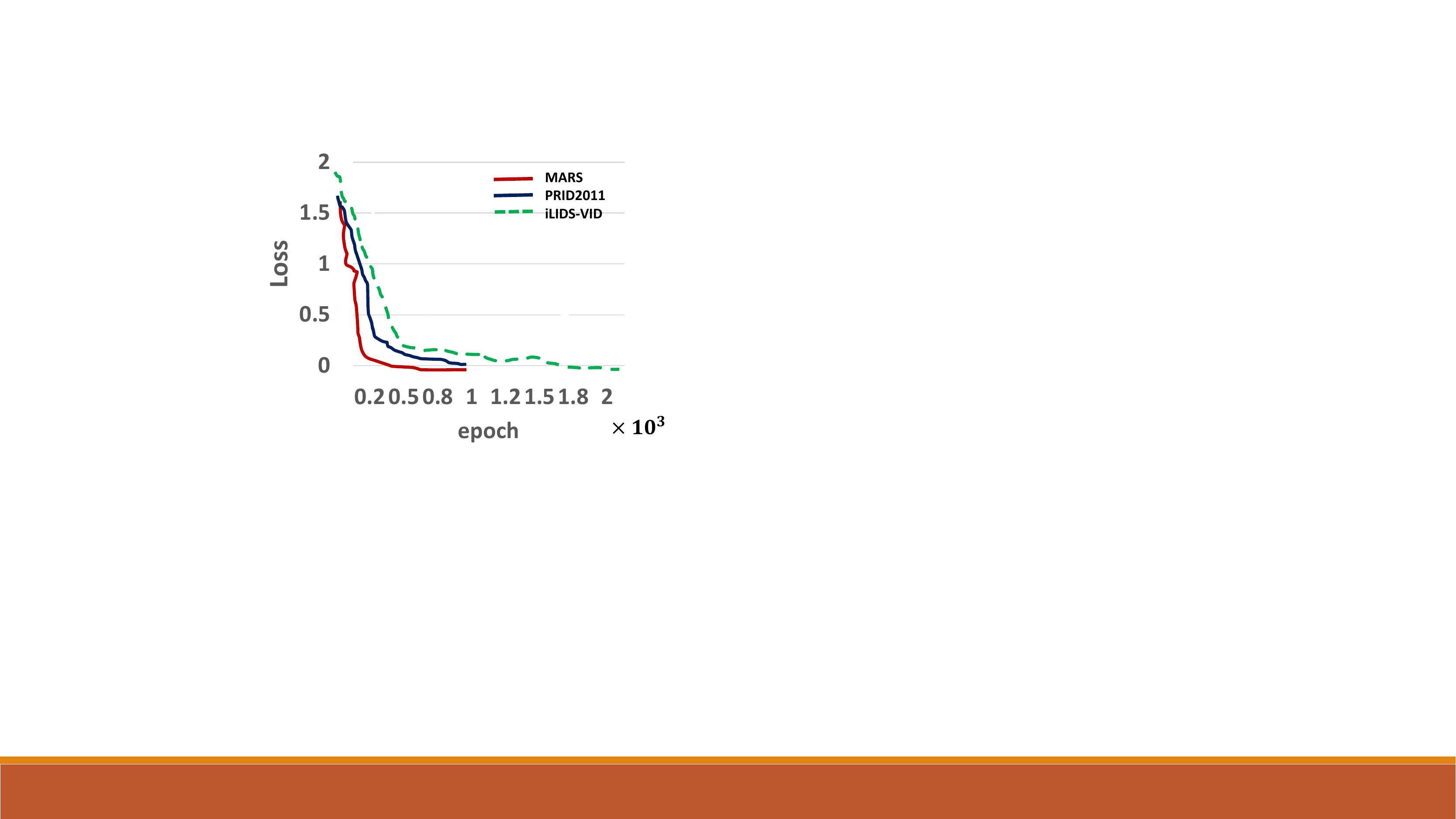}\\
(a) & (b)
\end{tabular}
\caption{(a) The loss is computed from the cross-validation (90\% of training split as training against 10\% of training split as validation set) w.r.t the varied value of $\lambda$ on the MARS dataset. (b) The convergence study on three datasets with increased training epoches.}
\label{fig:study-parameter}
\end{figure}

Since our network is Siamese alike, we randomly select positive and negative sequence pairs in on-line manner. Nonetheless, each positive/negative sequence pair consists of an arbitrary length in their full sequences to depict the same person or different persons under disjoint cameras. Thus, to ensure the fairness of experiments, we follow the same sequence length setting in \cite{RCNRe-id}. During training, sub-sequences of $T$=20 consecutive frames are used for computational purpose, where a different subset of 20 frames are randomly selected from the whole video sequence at each epoch. The network is trained with 1,000 epochs on the iLIDS-VID and PRID 2011 datasets while 2,000 epochs on the MARS dataset (The convergence evaluations are shown in Fig. \ref{fig:study-parameter} (b). It takes 0.06s for a pair of frames on a single NVIDIA GeForce GTX 980 GPU with 12GB memory.). In testing, we regard the first camera as the probe while the second camera as the gallery.

We did self data augmentation in training by cropping and mirroring all frames for a given sequence. In addition, we artificially augment the data by performing random 2D translation \cite{RCNRe-id}. For a frame of size $A\times B$, the same sized frames around the image center are sampled with translation drawn from a uniform distribution in the range of $[-0.05A, 0.05A] \times [-0.05B, 0.05B]$. In testing, data augmentation is also applied into the probe and the gallery sequences in the case that the length of each person sequence is less than 128, and the similarity scores between sequences are computed and averaged over all augmentation conditions \cite{AugmentationREID}.

In our deep GRU, all input-to-hidden and hidden-to-hidden kernels are of size $5\times 5$, \ie $k_1=k_2=5$. The structure setting is optimized by extensive self-study on architecture (Please refer to Table \ref{tab:detail_arch}). We apply zero-padded $5\times 5$ convolutions on each ConvGRU to preserve the spatial dimension. The stacked GRU consists of 3 layers with 128, 256, 256 channels, respectively. Max-pooling is applied on hidden-representations between the recurrent layers for the compatibility of the spatial dimensions. The hidden representations at the top of GRU across frames are average pooled to be the overall representation for the complete sequence. In our training, the convolutional activations with parameters regarding GoogLeNet are pre-trained on ImageNet, while the parameters of ConvGRUs are initialized by sampling them from an uniform distribution within an interval [-1, 1]. The whole set of parameters are trained/updated by using a RMSProp \cite{RMSProp} stochastic gradient descent procedure with mini-batches of 10 pairs of time series. To prevent the gradients from exploding, we clip all gradients to lie in the interval [-5, 5] \cite{AdvanceRNN}. A dropout ratio of 0.7 is applied on the hidden-unit activations. We define similar video sequence to be those from the same person and dissimilar ones to be those from different persons. For the penalty coefficient, we study the effect with varied values (see Fig. \ref{fig:study-parameter} (a)) and finally we experimented with $\lambda=1$ in all experiments.

\subsection{Baselines and Evaluation Metric}

We consider a variety of baselines in our experiments:
\begin{itemize}
\item (1) VGG 16 + FC-GRU: A deep architecture with FC-GRUs, which are based on VGG 16 model \cite{VGG}, pre-trained on ImageNet \cite{ImageNet}, and fine-tune it on the MARS dataset. We extract features from fully-connected layer $fc7$ (This can be viewed as a feature map with $1\times1$ spatial dimension), which are given as the inputs to the proposed deep network. This baseline has three layers of GRUs with 128, 256, and 256 channels, respectively. And the kernel sizes across three layers are set to be $k_1=k_2=1$ throughout.

\item (2) GoogLeNet + FC-GRU: The architecture structure is the same as VGG16 + FC-GRU except that the features are extracted from the last fully connected layers of Inception module.

\item (3) $Ours_{G}$ ($9\times 9, 1\times 1$): In all variants of our model, the input-to-state kernel size in the first layer GRU is set to be $3 \times 3$ (\ie $\bx_t=\bh_t^0 \xrightarrow[]{3\times 3} \bh_t^1$) while the state-to-state kernel sizes across the second and third layers are varied to determine the variants of each model. We modify the network with the state-to-state kernels of size $9\times 9$ and $1\times 1$ for the second and third layers of GRU, respectively. The features are extracted from the last convolutional layer of GoogLeNet. Table \ref{tab:detail_arch} shows the configurations of all variants of our model.

\item (4) $Ours_{G}$ ($1\times 1, 9\times 9$): This variant has the same configuration as $Ours_{G}$ ($9\times 9, 1\times 1$) while the sizes of state-to-state kernels for the second and third layers of GRU are set to be $1\times 1$ and $9\times 9$, respectively.

\item (5) $Ours_{G}$ ($5\times 5, 5\times 5$): The configuration is the same as above while the sizes of state-to-state kernels for the second and third layers of GRU are set to be both $5\times 5$.

\item (6) $Ours_{V}$ ($9\times 9, 1\times 1$): This variant has the same configuration as $Ours_{G}$ ($9\times 9, 1\times 1$) while the features are extracted from the last convolution of VGG-16.

\item (7) $Ours_{V}$ ($1\times 1, 9\times 9$): This variant has the same configuration as $Ours_{G}$ ($1\times 1, 9\times 9$) while the features are extracted from the last convolution of VGG-16.

\item (8) $Ours_{V}$ ($5\times 5, 5\times 5$): This variant has the same configuration as $Ours_{G}$ ($5\times 5, 5\times 5$) while the features are extracted from the last convolution of VGG-16.

\item (9) $\widehat{Ours_{G}}$ ($5\times 5, 5\times 5$): This variant has the same configuration as $Ours_{G}$ ($5\times 5, 5\times 5$) whereas the attention effect is muted by fixing all attention weights to an equal value of $1/K^2=1/49$.

\end{itemize}

The performance is evaluated by the average Cumulative Matching Characteristics (CMC) curves after 10 trials with different train/test split. Specifically, in the testing phase, the Euclidean distances between probe sequence features and those of gallery sequences are computed. Then, for each probe person, a rank order of all the candidates in the gallery is sorted from the one with the smallest distance to the largest distance. Finally, the percentage of true matches sorted out among the first $R$ ranked persons is computed and denoted as rank@$R$.

\subsection{Experimental Results}

\subsubsection{Study on Architecture Variations}
In this experiment, we study the properties of our model by comparing with variants in terms of varied convolutional kernel sizes: ($9\times 9$, $1\times1$), ($1\times 1$, $9\times9$), ($5\times 5$, $5\times5$), and VGG 16 /GoogLeNet + FC-GRU. Evaluation results are presented in Table \ref{tab:com_arch}. Our experiments show that the variants of deep attention with ConvGRUs perform consistently better than VGG-16 /GoogLeNet + FC-GRU. This is mainly because low-level percepts from convolution layers provide more spatial information in relation to a person's moving patterns whereas FC-GRUs ignore these spatial variations among time steps. More specific, the performances of $Ours_{\ast}$ ($9\times 9$, $1\times1$) are inferior to those with a larger state-to-state kernel size \ie $Ours_{\ast}$ ($1\times 1$, $9\times9$) and $Ours_{\ast}$ ($5\times 5$, $5\times5$). This provides evidence that larger state-to-state kernels are more suitable for capturing spatiotemporal correlations. Finally, additional contribution can be seen from the performance improvement caused by the attention mechanism where the rank-1 value drops when we remove attention in $\widehat{Ours_{G}}$ ($5\times 5, 5\times 5$).

\begin{table}[t]
\caption{Details on architecture variants. $\ast$ denotes G (GoogLeNet) or V (VGG 16).}\label{tab:detail_arch}
\centering
\begin{tabular}{r|r|r}
\hline
$Ours_{\ast}$ ($9\times 9, 1\times 1$) & $Ours_{\ast}$ ($1\times 1, 9\times 9$) & $Ours_{\ast}$ ($5\times 5, 5\times 5$) \\
\hline
$\bx_t=\bh_t^0 \xrightarrow[]{3\times 3} \bh_t^1$ & $\bx_t=\bh_t^0 \xrightarrow[]{3\times 3}  \bh_t^1$ & $\bx_t=\bh_t^0 \xrightarrow[]{3\times 3}  \bh_t^1$\\
$[\bh_t^1,\bh_{t-1}^2]\xrightarrow[]{9\times 9} \bh_t^2$ & $[\bh_t^1, \bh_{t-1}^2]\xrightarrow[]{1\times 1} \bh_t^2$ & $[\bh_t^1, \bh_{t-1}^2] \xrightarrow[]{5\times 5} \bh_t^2$\\
$[\bh_t^2,\bh_{t-1}^3 ]\xrightarrow[]{1\times 1} \bh_t^3$ & $[\bh_t^2,\bh_{t-1}^3 ] \xrightarrow[]{9\times 9} \bh_t^3$ & $[\bh_t^2,\bh_{t-1}^3 ] \xrightarrow[]{5\times 5} \bh_t^3$\\
\hline
\end{tabular}
\end{table}

\begin{table*}[t]
\caption{Evaluation on architecture variants.}\label{tab:com_arch}
\centering
\scriptsize
\begin{tabular}{l|c|c|c|c|c|c|c|c|c|c|c|c}
\hline
Dataset & \multicolumn{4}{c|}{iLIDS-VID} & \multicolumn{4}{c|}{PRID2011}  & \multicolumn{4}{c}{MARS}\\
\cline{1-13}
Rank @ R & R = 1 & R = 5 & R = 10 & R = 20 & R = 1 & R = 5 & R = 10 & R = 20 & R = 1 & R = 5 & R = 10 & R = 20\\
\hline
VGG 16 + FC-GRU &  52.6 & 72.5 & 81.3 & 88.5 & 62.2 & 81.8 & 84.2 & 89.7 & 56.7 & 73.4 & 79.8 & 84.4\\
GoogLeNet + FC-GRU & 53.8  & 73.2 & 83.0 &89.4  & 62.6 &83.0  & 85.0 & 91.2 &56.0 & 72.7& 79.1 & 84.0\\
$Ours_V$ ($9\times 9$, $1\times 1$)  &  54.5 & 73.8 &83.7 & 90.0 &63.0 & 85.0 & 86.5 & 92.5 & 57.6 & 74.8 & 80.7& 86.2\\
$Ours_V$ ($1\times 1$, $9\times 9$)  & 55.1 & 75.0 & 85.0 & 91.5 & 63.7 & 85.8 & 88.0 & 93.0  &57.8  &75.6 &81.5 &87.2\\
$Ours_V$ ($5\times 5$, $5\times 5$) & 56.7 & 77.0 & 87.5 & 92.5 & 65.2 &  88.0 & 89.0 & 93.5 &60.5 & 78.4& 82.0 & 89.0\\
\hline
$Ours_G$ ($9\times 9$, $1\times 1$)  & 56.1  & 76.7  & 86.1 & 93.5 & 68.1 & 89.2 & 94.5 & 96.1 & 63.4 & 79.2 & 82.5 & 89.0\\
$Ours_G$ ($1\times 1$, $9\times 9$)  & 57.3 & 78.2 & 86.8 & 94.0 & 69.5 & 91.0 & 94.9 & 96.7 & 64.0 & 79.7 & 83.0 & 89.8\\
$Ours_G$ ($5\times 5$, $5\times 5$) & \textbf{61.2} & \textbf{80.7} & \textbf{90.3} & \textbf{97.3} & \textbf{74.8} & \textbf{92.6} & \textbf{97.7} & \textbf{98.6} & \textbf{69.7} & \textbf{83.4} & \textbf{88.3} & \textbf{93.6}\\
$\widehat{Ours_{G}}$ ($5\times 5, 5\times 5$) &51.8  & 70.4 &  78.8 & 85.9 &58.6 & 77.7 & 80.2 & 86.8 & 55.2 & 70.6 & 77.1 & 82.3 \\
\hline
\end{tabular}
\end{table*}

\subsubsection{Study on Temporal Pooling with Attention}

To achieve a video-level representation where the local descriptor relies on recurrence at frame-level, one needs to apply normalization on frame descriptors and then average pooling of the normalized descriptors over time. In our approach, the average/attention pooling over frames is defined in Eq.\eqref{eq:temporal-selection}. An alternative is max pooling which can select the maximum activation of each element from frame-level features: $\bh_{video} = \max(\bh_1,\ldots, \bh_T)$. Fisher vector \cite{FisherVector} is a leading pooling technique which can provide state-of-the-art results in many different applications. In Fisher vector encoding, a Gaussian Mixture Model (GMM) with $C$ components can be denoted as $\Delta=\{(\mu_k,\sigma_k,\pi_k),k=1,2,\ldots,C\}$ where $\mu_k$, $\sigma_k$ and $\pi_k$ are the mean, variance and prior parameters of $c$-th component learned from the deep recurrent architecture in the frame level, respectively. Given $H=[\bh_1,\ldots, \bh_T]$ of deep descriptors extracted from a video by a network, the mean and covariance deviation vectors for the $c$-th component can be computed as:
\begin{equation}\scriptsize
\begin{split}
\textbf{u}_k=\frac{1}{T \sqrt{\pi_k}}\sum_{t=1}^T q_{kt}\left(\frac{\bh_t-\mu_k}{\sigma_k}\right),
\textbf{v}_k=\frac{1}{T \sqrt{2\pi_k}} \sum_{t=1}^T q_{kt}\left[\left(\frac{\bh_t-\mu_k}{\sigma_k}\right)^2 - 1\right],
\end{split}
\end{equation}
where $q_{kt}$ is the posterior probability. The Fisher vector is formed by concatenating $\textbf{u}_k$ and $\textbf{v}_k$ of all the $C$ components. Thus, the dimension of the Fisher vector is $2DC$ where $D$ is the dimension of the hidden representation $\bh_t$.

\begin{figure}[t]
\centering
\includegraphics[height=3.5cm]{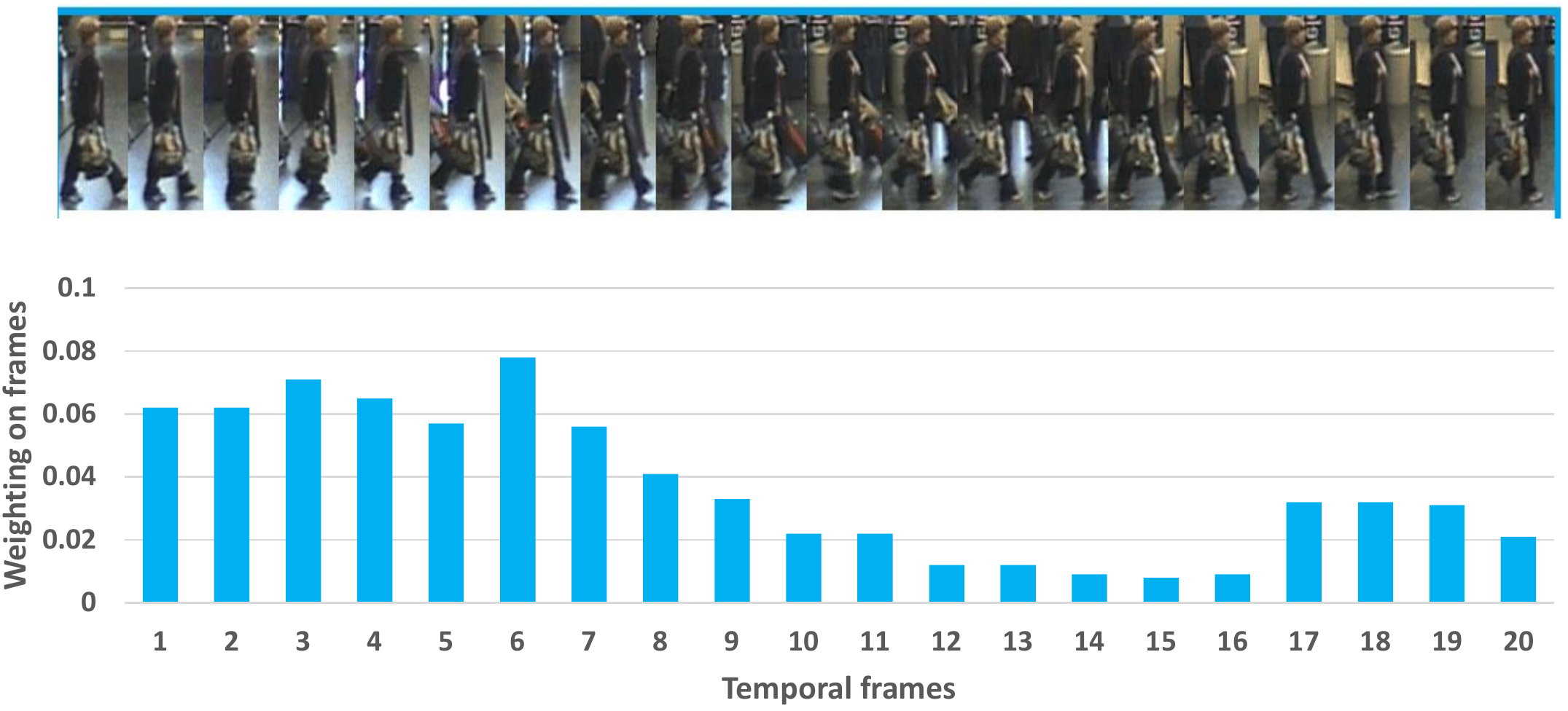}
\caption{When to focus: Temporal selection with soft attention over frames.}
\label{fig:soft-attention-temporal}
\end{figure}

We compare the performance using different pooling methods. Results are given in Table \ref{tab:com_pooling}. It can be seen that our average/attention pooling is superior to the max pooling. One reason is the average/attention pooling considers all the time steps equally important while subject to attention selection with re-weighting on each frame, whilst max pooling only employs the feature value in the temporal step with the largest activation. The illustration on temporal selection over informative frames is shown in Fig. \ref{fig:soft-attention-temporal}, which demonstrates that important frames are attached with higher weights while less important frames are eliminated. In Table \ref{tab:com_pooling}, Fisher vector encoding performs slightly better than our averaging/attention pooling. This is mainly because Fisher vectors use higher order statistics to model feature elements in a generative process. However, our average/attenion is advantageous in terms of end-to-end trainable and efficient in pooling whereas Fisher vector has high computational cost in updating parameters.

\begin{table*}[t]
\caption{Evaluation on different feature pooling schemes on top of convolutional GRUs.}\label{tab:com_pooling}
\centering
\scriptsize
\begin{tabular}{l|c|c|c|c|c|c|c|c|c|c|c|c}
\hline
Dataset & \multicolumn{4}{c|}{iLIDS-VID} & \multicolumn{4}{c|}{PRID2011} & \multicolumn{4}{c}{MARS}\\
\cline{1-13}
Rank @ R & R = 1 & R = 5 & R = 10 & R = 20 & R = 1 & R = 5 & R = 10 & R = 20 & R = 1 & R = 5 & R = 10 & R = 20\\
\hline
Max pooling & 59.4 & 58.6 & 83.8  &  90.4 & 72.4  & 90.8 & 96.0 & 96.2 & 66.8 & 81.1 & 86.7 & 91.2\\
Average/attention pooling & 61.2 & 80.7 & 90.3 &97.3 & 74.8 & 92.6 & \textbf{97.7} & \textbf{98.6} & 69.7 & 83.4& 88.3 & \textbf{93.6}\\
Fisher vector pooling & \textbf{63.2} & \textbf{80.9} & \textbf{92.2}  & \textbf{97.5} & \textbf{79.2} & \textbf{92.7} & 97.4 & 98.1 & \textbf{70.4} & \textbf{85.0} & \textbf{89.6} & 93.2\\
\hline
\end{tabular}
\end{table*}

\subsubsection{Study on Varied Sequence Lengths}

\begin{figure}[t]
\begin{tabular}{cc}
\hspace{-0.5cm}\includegraphics[height=3.5cm]{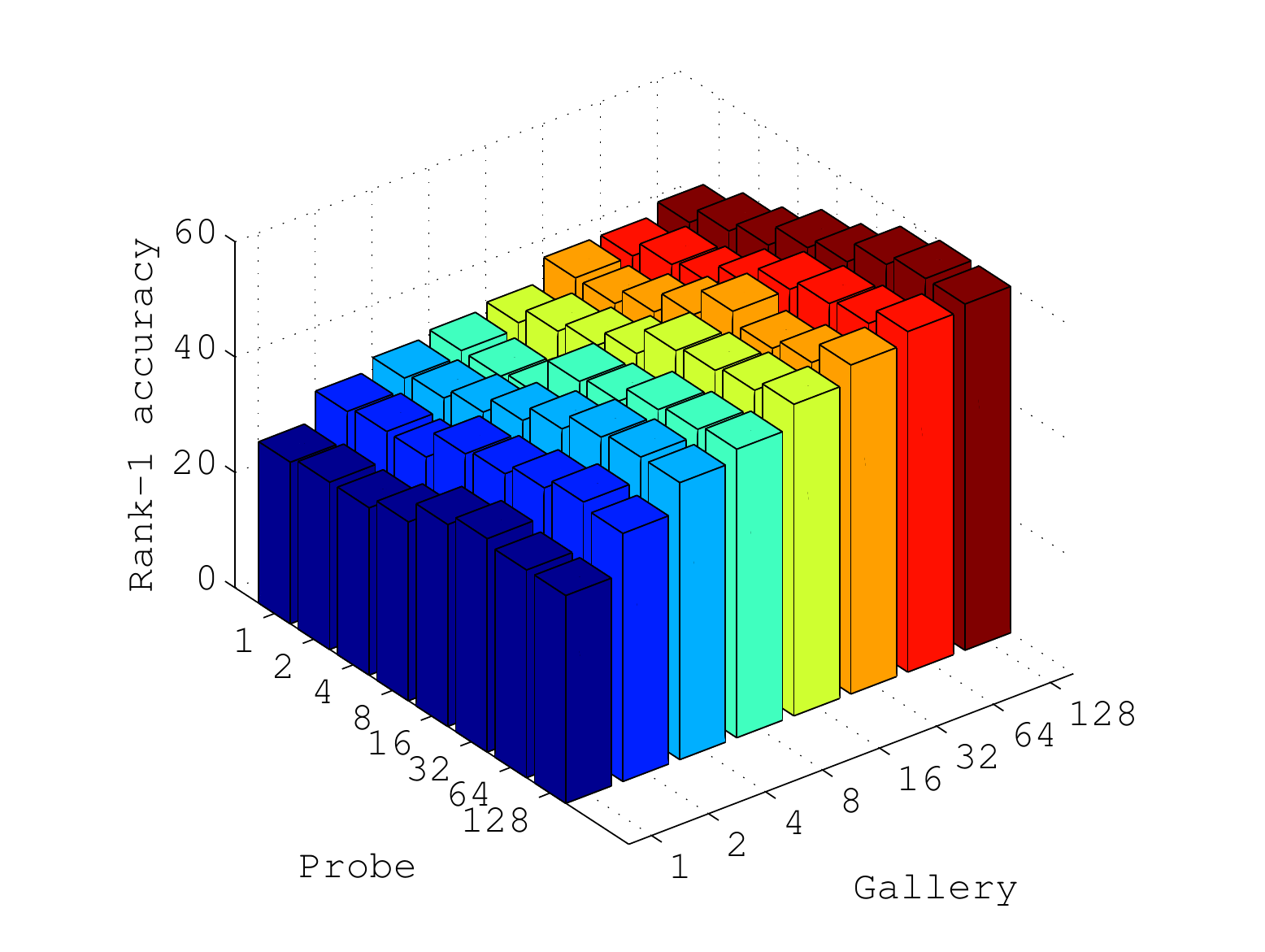}&
\hspace{-0.5cm}\includegraphics[height=3.5cm]{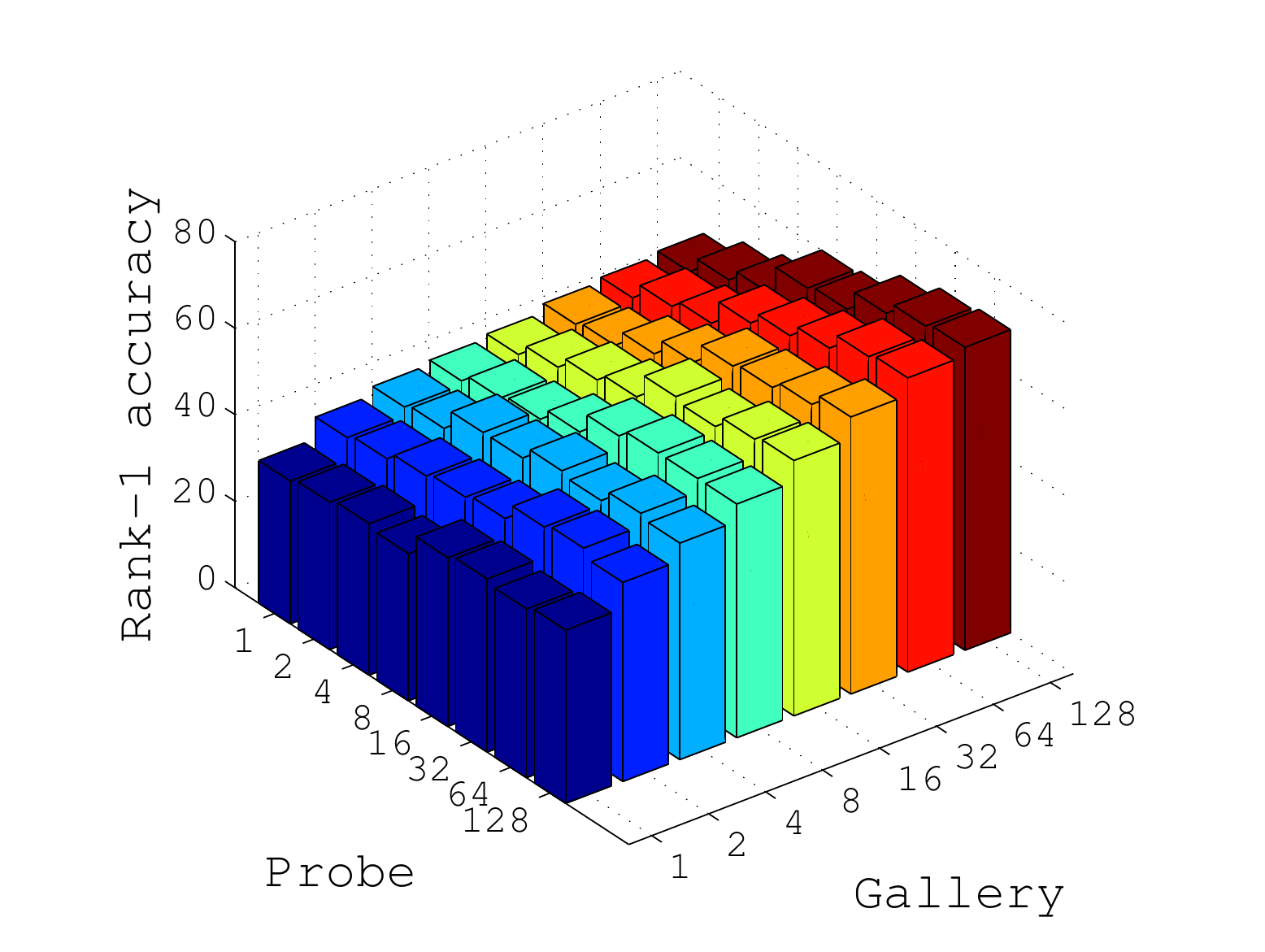}\\
(a) iLIDS-VID & (b) PRID2011
\end{tabular}
\caption{Rank-1 matching rate as varied length of probe and gallery sequences.}
\label{fig:varied_length}
\end{figure}

In this experiment, we investigate how the performance varies against length of the probe and gallery sequences in test stage. Evaluations are conducted on two datasets, and the lengths of the probe and gallery sequences are varied between 1 and 128 frames in step of power of two. Results are shown in Fig.\ref{fig:varied_length} where a bar matrix shows the rank-1 re-identification matching rate as a function of the probe and gallery sequence lengths. We can see that increasing the length in either probe or gallery can increase the matching accuracy. Also longer gallery sequences can bring about more benefits than longer probe sequences.

\subsubsection{Attention over Time}

\begin{figure}[t]
\centering
\includegraphics[height=3cm,width=9cm]{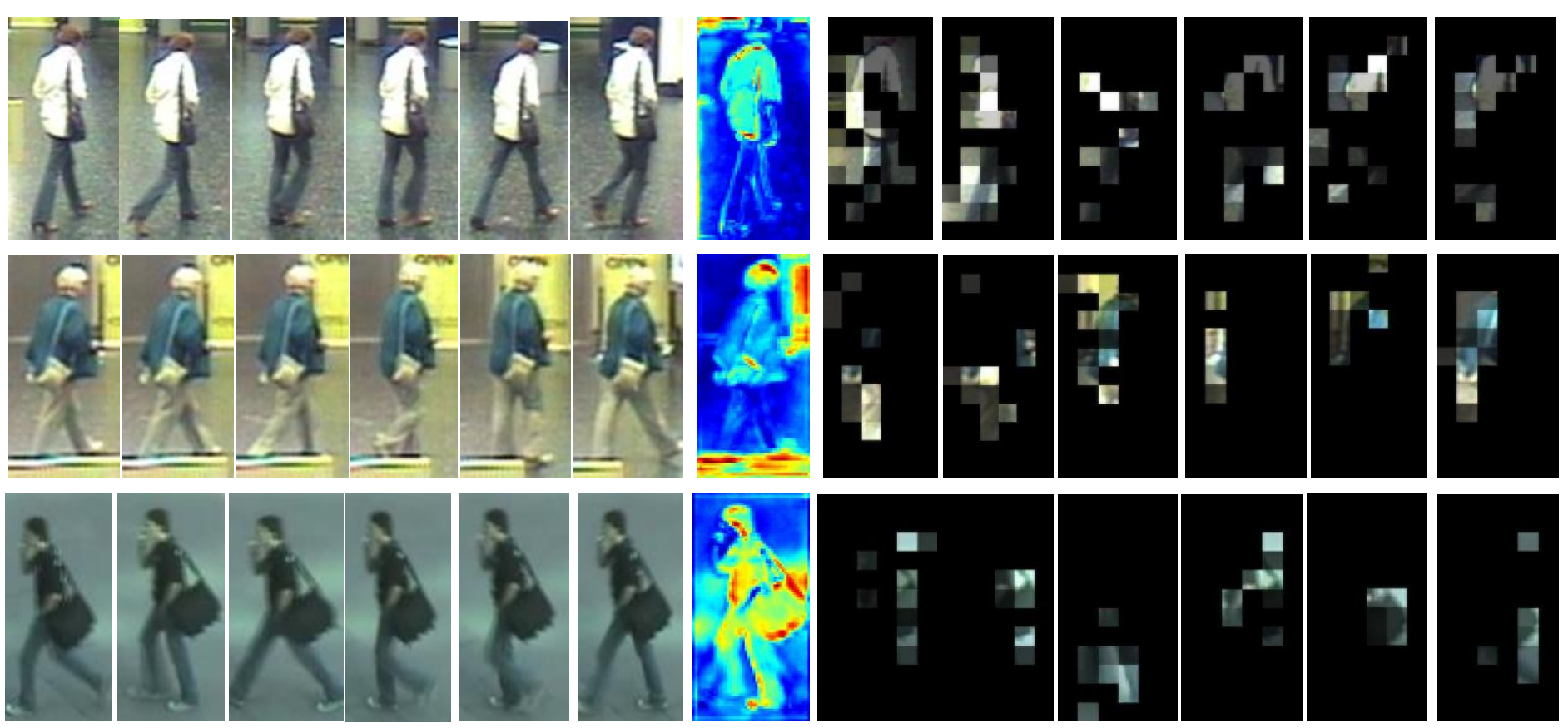}
\caption{Attention over time on different persons from iLIDS-VID (the upper two rows) and PRID2011 (the last row). Each row shows six original frames, a high-response feature map, and corresponding attention maps. Our method embeds attention into spatial local correlation to focus on distinct parts that help recognize the person in dramatic visual changes.}
\label{fig:attention-visualization}
\end{figure}

Fig. \ref{fig:attention-visualization} shows test examples of where our model attends on the iLIDS-VID and PRID2011 datasets. We can see that the model is able to focus on distinctive parts of the person and staying on them over time. For instance, on the cases of two persons from the iLIDS-VID dataset where the cluttered backgrounds render the recognition very challenging, our method is seen by effectively localizing the spatial regions, such as the jackets and the bags, which can help recognize the person. It can also be observed that the model is able to continuously attend to important parts in videos by focusing on some noticeable part \eg the large black bag (as shown in the last row). This can demonstrate the improved discriminative capability of local features produced by our deep attention Siamese networks. Hence, it shows that our attention model is capable of focusing on the distinct parts across time steps and capture the dynamic appearance of different people both spatially and temporally even in the context of varied background clutters and occlusions.

\subsection{Comparison with Other Representations}

\begin{table*}[t]
\caption{Comparison with different feature representations.}\label{tab:com_feature}
\centering
\scriptsize
\begin{tabular}{l|c|c|c|c|c|c|c|c|c|c|c|c}
\hline
Dataset & \multicolumn{4}{c|}{iLIDS-VID} & \multicolumn{4}{c|}{PRID2011} & \multicolumn{4}{c}{MARS}\\
\cline{1-13}
Rank @ R & R = 1 & R = 5 & R = 10 & R = 20 & R = 1 & R = 5 & R = 10 & R = 20& R = 1 & R = 5 & R = 10 & R = 20\\
\hline
HOG3D \cite{HOG3D}  & 8.3 & 28.7 & 38.3 & 60.7 & 20.7 & 44.5 & 57.1 & 76.8 & 2.6 & 9.9 & 12.4 & 20.4\\
FV2D  \cite{LocalFV} & 18.2 & 35.6 & 49.2 & 63.8 & 33.6 & 64.0 & 76.3 & 86.0 & 9.7 & 19.8 & 33.5& 43.5\\
FV3D \cite{VideoPerson} & 25.3 & 54.0 & 68.3 & 87.3 & 38.7 & 71.0 & 80.6 & 90.3 & 13.2& 24.3 & 38.8 & 52.4\\
STFV3D \cite{VideoPerson} & 37.0 & 64.3 & 77.0 & 86.9 & 42.1 & 71.9 & 84.4 & 91.6 & 15.4& 28.0& 40.3& 55.0\\
Ours & \textbf{61.2} & \textbf{80.7} & \textbf{90.3} & \textbf{97.3} & \textbf{74.8} & \textbf{92.6} & \textbf{97.7} & \textbf{98.6} & \textbf{69.7} & \textbf{83.4} & \textbf{88.3} & \textbf{96.6}\\
\hline
\end{tabular}
\end{table*}

In this experiment, we compare the learned hidden representations with four competing representations for person re-id.
\begin{itemize}
\item \textbf{FV2D} is a multi-shot approach \cite{LocalFV} which treats the video sequence as multiple independent images and uses Fisher vectors as features.

\item \textbf{HOG3D} extracts 3D HOG features \cite{HOG3D} from volumes of video data \cite{VideoRanking}. Specifically, after extracting a walk cycle by computing local maxima/minima of the FEP signal, video fragments are further divided into $2\times 5$ (spatial) $\times$ 2 (temporal) cells with 50\% overlap. A spatiotemporal gradient histogram is computed for each cell which is concatenated to form the HOG3D descriptor.

\item \textbf{FV3D} is similar to HOG3D where a local histogram of gradients is extracted from divided regular grids on the volume. However, we encode these local HOG features with Fisher vectors instead of simply concatenating them.

\item \textbf{STFV3D} is a low-level feature-based Fisher vector learning and extraction method which is applied to spatially and temporally aligned video fragments \cite{VideoPerson}. STFV3D proceeds as follows: 1) temporal segments are obtained separately by extracting walk cycles \cite{VideoRanking}, and spatial alignment is implemented by detecting spatial bounding boxes corresponding to six human body parts; 2) Fisher vectors are constructed from low-level feature descriptors on those body-action parts.
\end{itemize}

Experimental results are shown in Table \ref{tab:com_feature}. From the results, we can observe that our deep representation outperforms consistently over other representations. More specifically, HOG3D is inferior to Fisher vectors based features since Fisher vectors encode local descriptors in a higher order and suitable for person re-identification problem. It is not a surprise to see our features are superior to STFV3D because our deep networks work well in reconstructing spatiotemporal patterns.

\subsection{Comparison with Multi-shot Methods}

To evaluate the effectiveness of our model in producing deep spatiotemporal feature for person re-id, we compare the proposed method with a variety of multi-shot methods where multiple images of a sequence are exploited to construct more reliable descriptors to capture appearance variability. The competitors include

\begin{itemize}
\item SDALF \cite{Farenzena2010Person}: For each subject, local features including HSV histograms and colors are accumulated and weighted to form comparable features.
\item eSDC \cite{eSDC}: Patch-level dense distinctive features.
\item RankSVM \cite{RankSVM}+(Color,LBP): Color and Local Binary Pattern features are combined and averaged over each frame, then RankSVM is employed as distance metric.
\item PaMM \cite{PaMM}: This Pose-aware Multi-shot Matching robustly estimates poses and conducts multi-shot matching.
\item RFA-net \cite{RFA-net}: A progressive fusion framework based on LSTM aggregates the frame-wise human region representation and yields a sequence level feature representation.
\item PAM \cite{Multi-shot-WACV}: It produces a signature representation, namely Part Appearance Mixture by modeling a person's appearance as a multi-channel appearance mixture, where each channel corresponds to a particular region of the body.
\item S-LSTM \cite{S-LSTM}: For each fragment, the LSTM model processes image regions sequentially to obtain its frame-level representations, and then all frame-level features are aggregated into the video-level by using average pooling.
\end{itemize}

Table \ref{tab:multi_shot} shows the comparison results. It is evident that deeply learned space-time features produced by our deep attention model are more effective in matching person in multi-shot setting compared with low-level feature combination in SDALF \cite{Farenzena2010Person}, eSDC \cite{eSDC}, RankSVM \cite{RankSVM}, and PaMM \cite{PaMM}. This can be explained by the stacked convolution operations that extract local features robustly. Moreover, convolutional GRUs with attention selection can leverage spatial contexts as well as capture input signal variation in temporal dimension. Compared with the S-LSTM \cite{S-LSTM} that applies LSTM unit to learn deep yet context-aware features within each frame, our method outperform S-LSTM \cite{S-LSTM} in rank-1 accuracy by 8.8, 10.9, and 12.1 on three datasets, respectively.

\begin{table*}[t]
\caption{Comparison with multi-shot methods.}\label{tab:multi_shot}
\centering
\scriptsize
\begin{tabular}{l|c|c|c|c|c|c|c|c|c|c|c|c}
\hline
Dataset & \multicolumn{4}{c|}{iLIDS-VID} & \multicolumn{4}{c|}{PRID2011} & \multicolumn{4}{c}{MARS}\\
\cline{1-13}
Rank @ R & R = 1 & R = 5 & R = 10 & R = 20 & R = 1 & R = 5 & R = 10 & R = 20 & R = 1 & R = 5 & R = 10 & R = 20\\
\hline
SDALF \cite{Farenzena2010Person}  & 6.3 & 18.8 & 27.1 & 37.3 & 5.2 & 20.7 & 32.0 & 47.9 & 8.9 &32.1 & 40.3 & 56.7 \\
eSDC \cite{eSDC} & 10.2 & 24.8 &35.5 & 52.9 & 25.8 &43.6  & 52.6 & 62.0 &-&-&-&-\\
RankSVM \cite{RankSVM} & 23.2 & 44.2 & 54.1 & 68.8 & 34.3 & 56.0 & 65.5 & 77.3 &-&-&-&-\\
RFA-net \cite{RFA-net} & 49.3 & 76.8 & 85.3 & 90.0 & 58.2 & 85.8 & 93.4 & 97.9 &-&-&-&-\\
PaMM \cite{PaMM} & 30.3 & 56.3 & 70.3 & 82.7 & 56.5 & 85.7& 96.3 & 97.0 &-&-&-&-\\
PAM \cite{Multi-shot-WACV} & 33.3 & 57.8 & 68.5 &80.5 & 70.6 & 90.2 & 94.6 &97.1  &-&-&-&-\\
S-LSTM \cite{S-LSTM} & 52.4 & 74.2 & 86.5 & 92.1  & 63.9 &87.8 & 94.8 &97.4 & 57.6 &79.3 & 84.1&93.7 \\
Ours & \textbf{61.2} & \textbf{80.7} & \textbf{90.3} & \textbf{97.3} & \textbf{74.8} & \textbf{92.6} & \textbf{97.7} & \textbf{98.6} & \textbf{69.7} & \textbf{83.4} & \textbf{88.3} & \textbf{96.6}\\
\hline
\end{tabular}
\end{table*}

\subsection{Comparison with State-of-the-art Approaches}\label{ssec:state-of-the-art}

In this section, we compare our method with state-of-the-art approaches. Recent works \cite{VideoPerson,Video-person-matching} show that spatiotemporal features from sequences can be combined with distance metric learning algorithms to achieve better performance. Thus, we combine the learned deep features with two supervised metric learning methods: Local Fisher Discriminant Analysis (LFDA \cite{Pedagadi2013Local}) and KISSME \cite{KISSME}. The two methods are applied on top of deep features. Specifically, in the two methods, PCA is first performed to reduce the dimensionality of the original representation, and the reduced dimension is empirically chosen to be 150, as suggested by \cite{VideoPerson}.

\begin{figure}[t]
\begin{tabular}{cc}
\hspace{-0.5cm}\includegraphics[height=4cm]{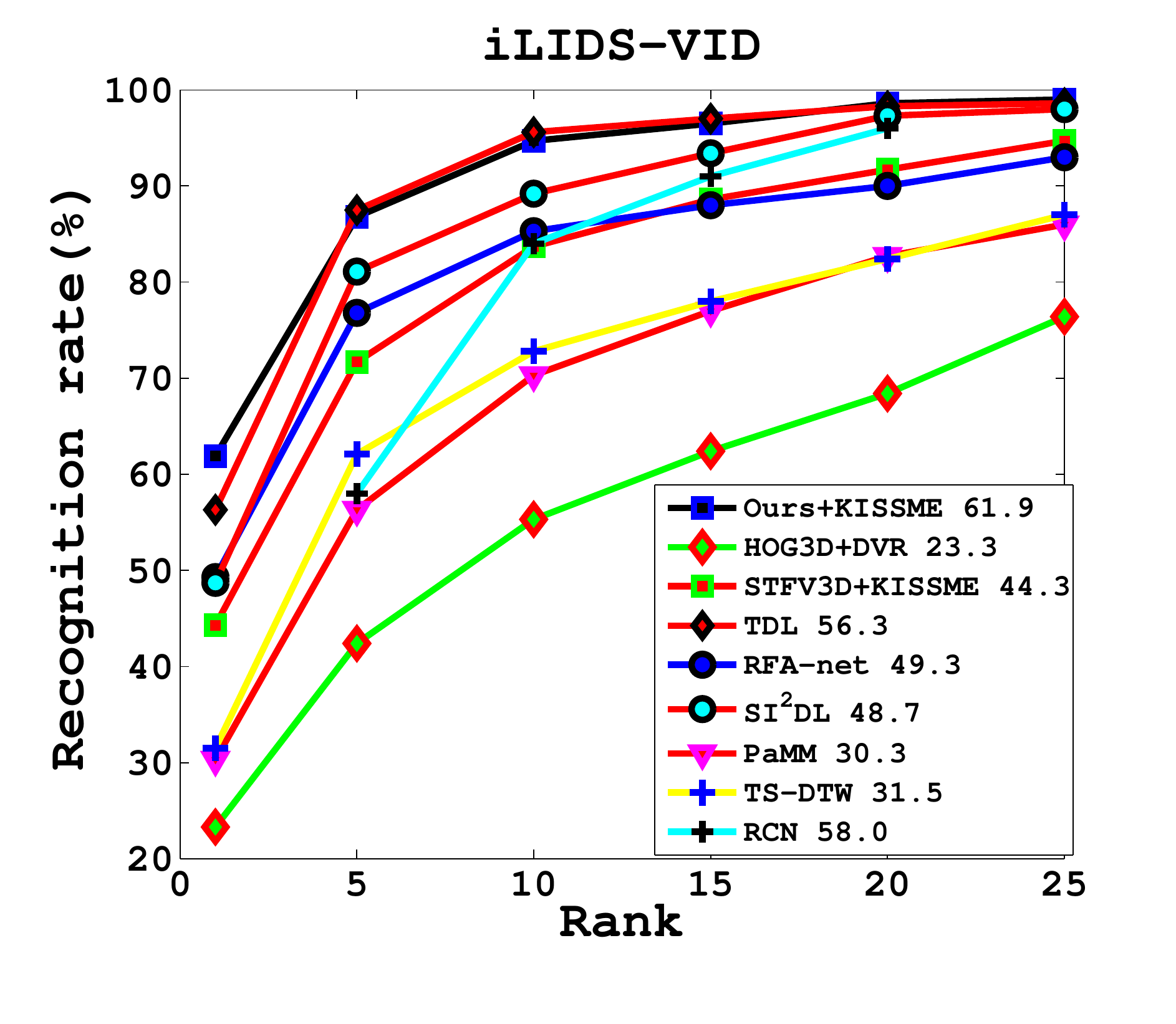}&
\hspace{-0.5cm}\includegraphics[height=4cm]{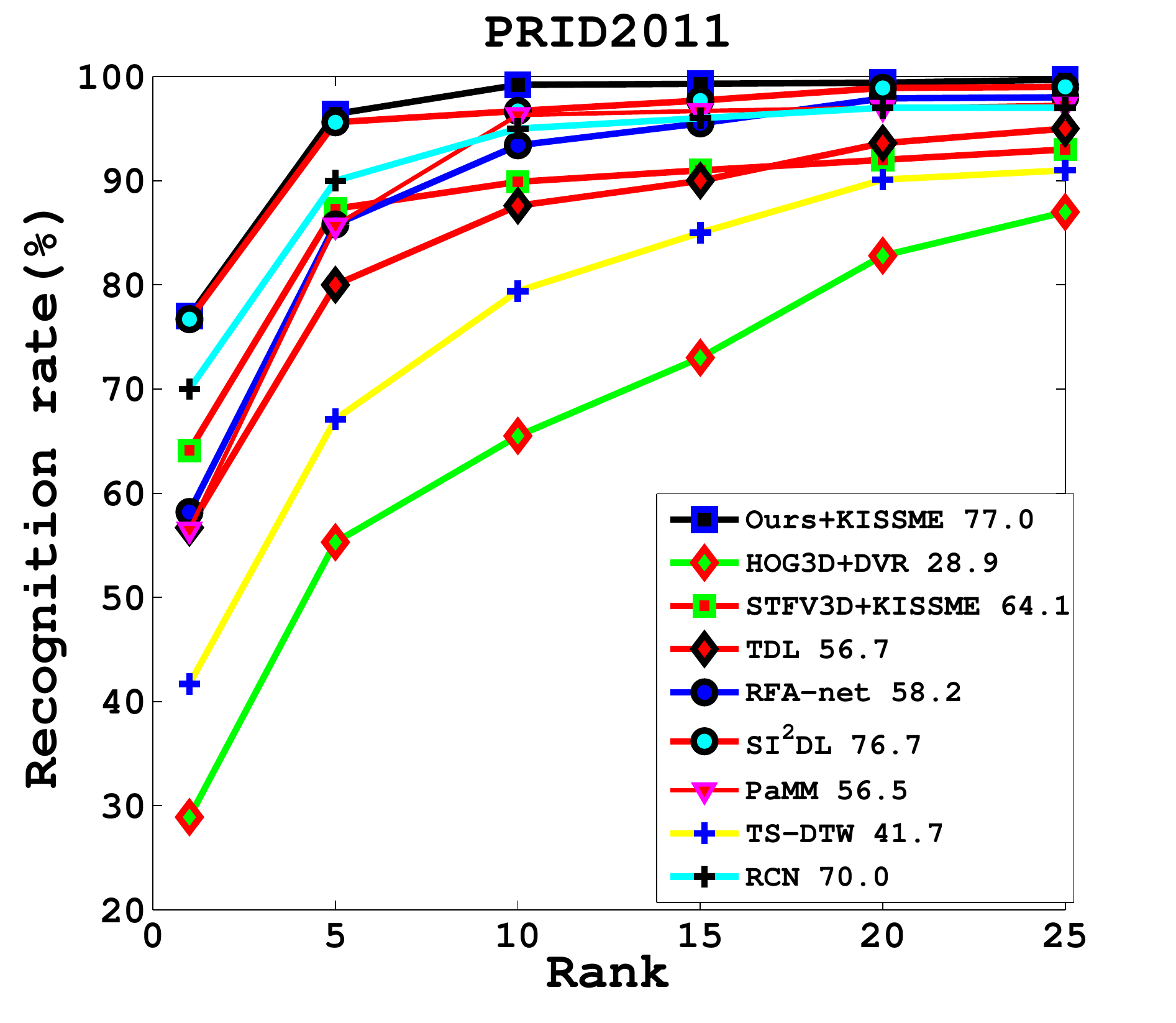}\\
(a) iLIDS-VID &  (b) PRID2011
\end{tabular}\caption{CMC curves on the iLIDS-VID and PRID2011 datasets. Rank-1 matching rate is marked after the name of each approach.}\label{fig:match_rate}
\end{figure}

In this experiment, we consider the competitors as follows: RCN \cite{RCNRe-id}, HOG3D+Discriminative Video Ranking (DVR) \cite{VideoRanking}, TDL \cite{Top-push}, RFA-net \cite{RFA-net}, $SI^2DL$ \cite{Video-person-ijcai16}, STFV3D \cite{VideoPerson}+ LFDA \cite{Pedagadi2013Local}, and STFV3D \cite{VideoPerson}+ KISSME \cite{KISSME}. Table \ref{tab:com_state} and Fig.\ref{fig:match_rate} show the comparison results. It is seen that distance metric learning can further improve the performance of our method. Notably our method combined with KISSME achieves rank-1 accuracy of $61.9\%$, $77.0\%$ and $73.5\%$ on the iLIDS-VID, PRID2011 and MARS datasets, outperforming all benchmark methods. In particular, comparing with TDL \cite{Top-push} that optimizes a distance learning method based on the learning to rank principle, our model can jointly learn video features and its similarity metric. RFA-net \cite{RFA-net} also employs LSTM to capture long-range dependencies to fuse frame-level features into video-level representations, our deep model is advantageous by leveraging spatial priors to improve the discriminative capability of local features.

\begin{table*}[t]
\caption{Comparison with state-of-the-art methods.}\label{tab:com_state}
\centering
\scriptsize
\begin{tabular}{l|c|c|c|c|c|c|c|c|c|c|c|c}
\hline
Dataset & \multicolumn{4}{c|}{iLIDS-VID} & \multicolumn{4}{c|}{PRID2011} & \multicolumn{4}{c}{MARS}\\
\cline{1-13}
Rank @ R & R = 1 & R = 5 & R = 10 & R = 20 & R = 1 & R = 5 & R = 10 & R = 20 & R = 1 & R = 5 & R = 10 & R = 20\\
\hline
HOG3D+DVR \cite{VideoRanking} & 23.3 & 42.4 & 55.3 & 68.4 & 28.9& 55.3 & 65.5 & 82.8 & 12.4 &33.2 & 54.7 &71.8\\
STFV3D \cite{VideoPerson} + LFDA \cite{Pedagadi2013Local} & 38.3 & 70.1 & 83.4 & 90.2 & 48.1 & 81.2 & 85.7 & 90.1 &20.1 & 31.2& 44.1& 57.7\\
STFV3D \cite{VideoPerson} + KISSME \cite{KISSME}& 44.3 & 71.7 & 83.7 & 91.7 & 64.1 & 87.3 & 89.9 & 92.0 & 22.0 & 33.4 & 44.8 & 59.0\\
TDL \cite{Top-push} & 56.3 &87.6 & 95.6 & 98.3 & 56.7 & 80.0 & 87.6 & 93.6 &-&-&-&-\\
RFA-net \cite{RFA-net} & 49.3 & 76.8 & 85.3 & 90.0 & 58.2 & 85.8 & 93.4 & 97.9 &-&-&-&-\\
$SI^2DL$ \cite{Video-person-ijcai16} & 48.7& 81.1 & 89.2 & 97.3 & 76.7 & 95.6 & 96.7 & 98.9 &-&-&-&-\\
PaMM \cite{PaMM} & 30.3 & 56.3 & 70.3 & 82.7 & 56.5 & 85.7& 96.3 & 97.0 &-&-&-&-\\
TS-DTW \cite{Video-person-matching} & 31.5 & 62.1 & 72.8 & 82.4 & 41.7 & 67.1& 79.4 & 90.1 &-&-&-&-\\
RCN \cite{RCNRe-id} & 58.0 & 84.0 & 91.0 & 96.0 & 70.0 & 90.0 & 95.0 & 97.0 &54.7&79.1&  83.6 &88.4\\
Ours & 61.2 & 80.7 & 90.3 & 97.3 & 74.8 & 92.6 & 97.7 & 98.6 & 69.7 & 83.4& 88.3 & 96.6\\
Ours + LFDA \cite{Pedagadi2013Local}& 61.4 & 83.4 & 91.0 & 97.9 & 75.7  &94.8 & 98.3 & 98.8 &71.3 & \textbf{85.1}& \textbf{89.5} & 96.9\\
Ours + KISSME \cite{KISSME} & \textbf{61.9} & \textbf{86.8} & 94.7 & \textbf{98.6} & \textbf{77.0} & \textbf{96.4} & \textbf{99.2} & \textbf{99.4} & \textbf{73.5} & 85.0 & \textbf{89.5} & \textbf{97.5} \\
\hline
\end{tabular}
\end{table*}

\subsection{Failure Examples}

\begin{figure}[t]
\centering
\includegraphics[height=3cm,width=7cm]{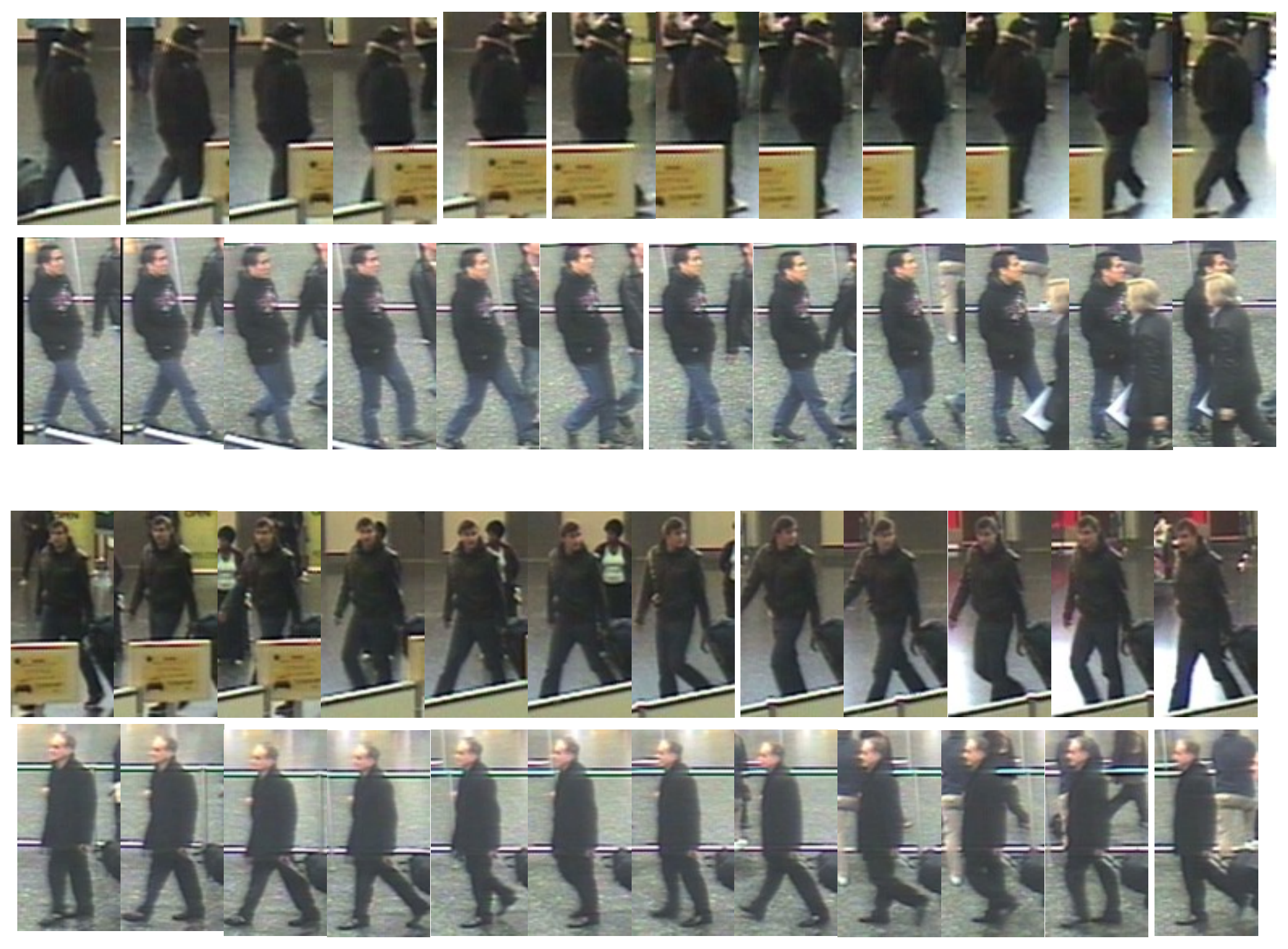}
\caption{The showcase of failure examples from the iLIDS-VID dataset.}
\label{fig:failure_example}
\end{figure}

In this experiment, we discuss the limitations of our approach, and show some failure examples on the iLIDS-VID dataset in Fig. \ref{fig:failure_example}. For each pair of video segments, the top row displays the query sequence and the bottom row shows the gallery sequence. It shows that our method is degenerated in the case when some different pedestrians exhibit very similar appearance in their upper bodies and walking gait patterns. For example, in the first matching case, the lower body parts of query are occluded by a notice sign. Even though our approach is more likely focusing on visual patterns on upper parts, as demonstrated in Fig. \ref{fig:attention-visualization}, the visual similarity of upper parts between the query person and the retrieved wrong person causes trouble for our method to identify the correct candidate from the gallery.

\subsection{Cross-Dataset Testing}\label{ssec:cross-dataset}
In this experiment, we perform cross-dataset testing to understand how well our approach generalizes across different training and testing datasets. This is a real-world problem due to the dataset bias \cite{AugmentationREID}, which is a form of over-fitting: the performance of a machine-learning based system, trained on that a particular dataset, is much worse when evaluated on a different dataset. Specifically, our deep model is trained on the largest and diverse MARS dataset and transferred to the target domains on 50\% of the iLIDS-VID and PRID 2011 datasets. The results of this experiment are reported in Table \ref{tab:cross_dataset}. In the cross-dataset scenario, recognition results are worse, as expected and probably due to the dataset bias. However, our framework's performance is not much below, and is well above other deep learning systems. We consider to improve the generalization performance of our re-identification system in future with some strategy in cross-domain learning.

\begin{table}[t]
\caption{Cross-dataset testing on iLIDS-VID and PRID2011. Training is performed on the MARS dataset.}\label{tab:cross_dataset}
\centering
\scriptsize
\begin{tabular}{|c|c|c|c|c|c|c|}
\hline
Dataset & \multicolumn{3}{c|}{iLIDS-VID} & \multicolumn{3}{c|}{PRID2011} \\
\cline{1-7}
Rank @ R & R = 1 & R = 10 & R = 20 & R = 1  & R = 10 & R = 20\\
\hline
Ours & \textbf{49.2} & \textbf{87.6} & \textbf{92.1} & \textbf{61.4} & \textbf{91.2} & \textbf{94.7}\\
RCN \cite{RCNRe-id} & 38.1 & 84.7 & 90.9 & 42.0 & 84.6 & 86.9\\
S-LSTM \cite{S-LSTM} & 30.6 & 74.8 & 83.3 & 37.0 & 81.4 & 84.7 \\
\hline
\end{tabular}
\end{table}

\section{Conclusion and Future Works}\label{sec:con}
In this paper, we present a deep attention based Siamese model to jointly learn spatiotemporal expressive video representations and similarity metrics for video-based person re-identification. Our approach embeds visual attention into convolutional activations from local regions to dynamically encode spatial context priors and capture the relevant patterns for the propagation through the network. As a consequence, local features are augmented to be highly discriminative to recognize persons in a challenging video scenario. Experimental results are conducted over three benchmark datasets to demonstrate the state-of-the-art performance of our method for video-based person re-id. In particular, we carefully study the structure and configuration of our network model by extensive self-evaluations on varied CNN models and spatial kernel sizes in stacked GRU layers. Also, some attention visualizations are present to show the effectiveness of our approach in combating against cluttered backgrounds and occlusions. Thus, our method is demonstrated to be effective in aligning the dynamic appearance of different people both spatially and temporally.

There are some interesting directions for further improvement of our framework. From the spatial alignment perspective, the arbitrary change of viewpoints still cause problems in spatial alignment. Thus, we are interested in investigating robust body part models to address the pose/viewpoint issue. The other direction is to explore a spatiotemporal primitive unit as the attention element, which can improve the generalization ability of our framework. We are also interested in developing an end-to-end trainable network that considers Fisher vector encoding as a pooling strategy to select discriminative frames amongst only a few video frames.

\section*{Acknowledge}

This work is partially supported by ARC DP 160104075. Junbin Gao's research is partially supported by Australian Research Council Discovery Projects funding scheme (Project No. DP140102270) and the University of Sydney Business School ARC Bridging Fund.

\ifCLASSOPTIONcaptionsoff
  \newpage
\fi



\bibliographystyle{IEEEtran}\small
\bibliography{allbib}

\end{document}